\documentclass[3p,authoryear]{elsarticle}
\usepackage{natbib}
\usepackage{fullpage}
\usepackage[english]{babel}
\usepackage[utf8]{inputenc}
\usepackage{amsfonts}
\usepackage{amsmath}
\usepackage{mathtools}
\usepackage{graphicx}
\usepackage{multirow,booktabs}
\usepackage{float,subcaption,overpic}
\usepackage[colorinlistoftodos]{todonotes}
\usepackage{rotating}
\usepackage{algorithm}
\usepackage[noend]{algpseudocode}
\usepackage{textcomp,xcolor}
\definecolor{MatlabCellColour}{RGB}{250,250,250}
\definecolor{MatPurp}{rgb}{.625,.1406,.9375}
\usepackage{listings}

\lstdefinestyle{customc}{
	belowcaptionskip=.25\baselineskip,
	breaklines=true,
	frame=L,
	xleftmargin=\parindent,
	language=Matlab,
	showstringspaces=false,
	basicstyle=\small\ttfamily,
	keywordstyle=\bfseries\color{white!30!black},
	identifierstyle=\color{blue},  
	commentstyle=\itshape\color{green!60!black},
	stringstyle=\color{MatPurp},
	backgroundcolor=\color{MatlabCellColour}
}

\DeclareMathOperator*{\argmax}{arg\,max}
\DeclareMathOperator*{\Max}{maximize}
\DeclareMathOperator*{\Min}{minimize}
\DeclareMathOperator{\trace}{tr}
\DeclareMathOperator{\vol}{vol}
\newcommand{\ba}{\mathbf{a}}
\newcommand{\be}{\mathbf{e}}

\newcommand{\bPhi}{\mathbf{\Phi}}

\newcommand{\bxi}{\boldsymbol{\xi}}
\newcommand{\bphi}{\boldsymbol{\phi}}
\newcommand{\bSigma}{\mathbf{\Sigma}}
\newcommand{\bTheta}{\mathbf{\Theta}}
\newcommand{\bA}{\mathbf{A}}

\newcommand{\bC}{\mathbf{C}}
\newcommand{\bL}{\mathbf{L}}
\newcommand{\bS}{\mathbf{S}}

\newcommand{\bQ}{\mathbf{Q}}
\newcommand{\bR}{\mathbf{R}}

\newcommand{\bV}{\mathbf{V}}

\newcommand{\bx}{\mathbf{x}}
\newcommand{\bX}{\mathbf{X}}
\newcommand{\bY}{\mathbf{Y}}
\newcommand{\by}{\mathbf{y}}

\newcommand{\reals}{\mathbb{R}}
\newcommand{\ie}{\textit{i.e.}}
\newcommand{\ind}{\gamma}
\newcommand{\Ind}{\boldsymbol{\gamma}}

\title{\textbf{Predicting shim gaps in aircraft assembly\\ with machine learning and sparse sensing}\vspace{-.05in}}

\author[uwamath]{Krithika Manohar\corref{cor}} \ead{kmanohar@uw.edu}
  \cortext[cor]{Corresponding author. \emph{E-mail:} kmanohar@uw.edu} 
    \author[boeing]{Thomas Hogan}
    \author[boeing]{Jim Buttrick}
    \author[uwise,uwme]{Ashis G. Banerjee}
  \author[uwamath]{J. Nathan Kutz}  
  \author[uwme]{and Steven L. Brunton\vspace{-.1in}}
  \address[uwamath]{Department of
    Applied Mathematics, University of Washington, Seattle, WA 98195,
    United States\vspace{-.05in}}
    \address[boeing]{The Boeing Company, Seattle, WA 98008\vspace{-.05in}}
    \address[uwise]{Department of Industrial \& Systems Engineering, University of Washington, Seattle, WA 98195,
    United States\vspace{-.05in}}    
    \address[uwme]{Department of
    Mechanical Engineering, University of Washington, Seattle, WA 98195,
    United States\vspace{-.35in}}
      
\date{}

\begin{document}

 \begin{abstract}
A modern aircraft may require on the order of thousands of custom shims to fill gaps between structural components in the airframe that arise due to manufacturing tolerances adding up across large structures. 
These shims, whether liquid or solid, are necessary to eliminate gaps, maintain structural performance, and minimize pull-down forces required to bring the aircraft into engineering nominal configuration for peak aerodynamic efficiency.  
Currently, gap filling is a time-consuming process, involving either expensive by-hand inspection or computations on vast quantities of measurement data from increasingly sophisticated metrology equipment.  
In either case, this amounts to significant delays in production, with much of the time being spent in the critical path of the aircraft assembly.  

In this work, we present an alternative strategy for predictive shimming, based on machine learning and sparse sensing to first learn gap distributions from historical data, and then design optimized sparse sensing strategies to streamline the collection and processing of data.  
This new approach is based on the assumption that patterns exist in shim distributions across aircraft, and that these patterns may be mined and used to reduce the burden of data collection and processing in future aircraft.  
Specifically, robust principal component analysis is used to extract low-dimensional patterns in the gap measurements while rejecting outliers.  
Next, optimized sparse sensors are obtained that are most informative about the dimensions of a new aircraft in these low-dimensional principal components.  
We demonstrate the success of the proposed approach, called PIXel Identification Despite Uncertainty in Sensor Technology (PIXI-DUST), on historical production data from 54 representative Boeing 
 commercial aircraft.  
Our algorithm successfully predicts $99\%$ of the shim gaps within the desired measurement tolerance using around $3\%$ of the laser scan points that are typically required; all results are rigorously cross-validated. 
\end{abstract}


\maketitle

\vspace{-.2in}
\section{Introduction}
\label{sec:introduction}
Advanced manufacturing is increasingly becoming a data rich endeavor, with big data analytics addressing critical challenges in high-tolerance assembly~\citep{muelaner2013achieving}, operation planning~\citep{lu1990combined}, quality control~\citep{milo2015new} and supply chains~\citep{lavalle2011big}.    
The broad applicability of data science in manufacturing is reviewed in~\cite{harding2006data,lee2013recent,lechevalier2014towards,esmaeilian2016evolution}.  
Machine learning is a particularly promising tool for extracting actionable patterns in vast quantities of high-dimensional data that are difficult to visualize and/or interpret.   
Examples of machine learning in manufacturing systems abound, for example using topological data analysis~\citep{guo2017identification}, deep learning~\citep{ren2017multi}, and genetic algorithms to evaluate form tolerances~\citep{sharma2000genetic}.  
Modern aircraft assembly is at the forefront of integrating big data into manufacturing, with advances in metrology accelerating aircraft manufacturing processes in recent years~\citep{muelaner2010design,jamshidi2010manufacturing,muelaner2011measurement,muelaner2011integrated,martin2011metrology,muelaner2013achieving,maropoulos2014new}, for example in large composite structures~\citep{muelaner2010design}, in fuselage skin panels~\citep{lowe2007automated}, and in the wing box~\citep{chouvion2011interface}.  

Aircraft are built to exceedingly high tolerances, with components sourced from around the globe.  
Even when parts are manufactured to specification, there may be significant gaps between structural components upon assembly.  
One of the most time-consuming and expensive efforts in part-to-part assembly is the shimming required to bring an aircraft into the engineering nominal shape.  
Historically, parts are dry-fit, gaps are measured, and then custom shims are manufactured and inserted, often involving disassembly and reassembly.  
Recent efforts to bypass this manual shimming have involved \emph{predictive shimming} based on measurement data of component parts~\citep{jamshidi2010manufacturing,muelaner2010design,muelaner2011measurement,chouvion2011interface,muelaner2011integrated,muelaner2013achieving}.  
There are several patents and papers describing methods of high-tolerance measurement and manufacturing required for predictive shimming~\citep{marsh2008laser,marsh2010method,chouvion2011interface,boyl2014digitally,vasquez2014systems,valenzuela2015systems,boyl2016methods,antolin2016end}.  
With increasingly sophisticated and expensive metrology equipment, vast quantities of point cloud and laser scan data is being acquired and the burden has shifted from time-consuming manual shimming to time-consuming computational processing for predictive shimming.  
In either case, this amounts to significant delays in production, with much of the time being spent in the critical path of assembly.  
Reducing the burden of data collection and processing, and ultimately reducing delays for optimized aircraft assembly, could have significant financial implications, on the order of billions of dollars a year.  

In this work, we present an alternative approach to predictive shimming based on machine learning and sparse sensing.  
Instead of measuring each component of a new aircraft in isolation, we leverage historical production data to learn patterns in the shim gap distributions.  
In particular, the robust principal component analysis (RPCA)~\citep{rpca} provides an estimate of the dominant principal components that is robust to outlier measurements.  
Robust statistical methods are critically important for evaluating real-world data, as advocated by John. W. Tukey in the earliest days of data science~\citep{Huber2002as,Donoho2015data}.  
RPCA is based on the computationally efficient singular value decomposition (SVD)~\citep{Golub1965siamb}, and yields the most correlated spatial structures in the aircraft measurements, identifying areas of high variance across different aircraft.  
Next, based on the robust principal components obtained from historical data, we design a small subset of key spatial measurement locations that best inform the shim gap distribution of a new aircraft.  
Our procedure, called PIXel Identification Despite Uncertainty in Sensor Technology (PIXI-DUST), is based on recent advances in convex optimization for sensor placement~\citep{Brunton2016siap,Manohar2017csm,Manohar2016jfs}.  
We demonstrate the success of PIXI-DUST on historical production data from $54$ Boeing 
 aircraft, predicting $99\%$ of the shim gaps within the desired measurement tolerance using approximately $3\%$ of the available laser scan data.  

Several groups have used dimensionality reduction, such as principal component analysis (PCA) to model deformation in geometric surfaces for improved fitting~\citep{camelio2002compliant,carnelio2006identification,zhang2006comparative,lindau2013statistical}.  
In the automotive industry, PCA was used in combination with sparse sensing to identify key measurement locations for the characterization of compliant part assembly~\citep{camelio2002compliant,carnelio2006identification}.  
Such dimensionality reduction methods have also been compared with sparsity promoting techniques for outlier rejection and minimal description of surfaces~\citep{zhang2006comparative}.  
However, this work is the first to combine RPCA with a scalable greedy sparse sensor optimization for robust predictive shimming in the aerospace industry.

\section{Background}\label{sec:background}
The results in this work combine robust dimensionality reduction and sparse sensor optimization algorithms to dramatically reduce the number of measurements required to shim a modern aircraft.  
This section provides a foundation for the methods that will be synthesized and applied throughout the paper.  

\subsection{Robust principal component analysis}
Least-squares regression is highly susceptible to outliers and corrupted data.  
Principal components analysis (PCA) suffers from the same weakness, making it \emph{fragile} with respect to outliers.  
To address this sensitivity, \cite{rpca} introduced a robust principal components analysis (RPCA) that decomposes a data matrix $\bX$ into a  low-rank matrix $\bL$ containing dominant coherent structures, and a sparse matrix $\bS$ containing outliers and corrupt data:
\begin{equation}
\bX = \bL + \bS.
\end{equation}
The principal components of $\bL$ are \emph{robust} to the outliers and corrupt data in $\bS$.  
The RPCA decomposition has tremendous applicability for many modern problems of interest, including video surveillance (where the background objects appear in $\bL$ and foreground objects appear in $\bS$), face recognition, natural language processing and latent semantic indexing, and matrix completion. 

Mathematically, the goal of RPCA is to find matrices $\bL$ and $\bS$ that satisfy
\begin{equation}
\min_{\bL,\bS}\text{rank}(\bL) + \|\bS\|_0 \text{    such that } \bL + \bS = \bX.\label{Eq:RPCA:L0}
\end{equation}
However, neither the $\text{rank}(\bL)$ nor the $\|\bS\|_0$ terms are convex, and this is not a scalable optimization problem.  
Similar to compressed sensing~\citep{donoho2006compressed,candes2006compressive}, it is possible to solve for the optimal $\bL$ and $\bS$ with \emph{high probability} using a convex relaxation of~\eqref{Eq:RPCA:L0}:
\begin{equation}
\min_{\bL,\bS}\|\bL\|_* + \lambda\|\bS\|_1 \text{    such that } \bL + \bS = \bX.\label{Eq:RPCA:L1}
\end{equation}
Here, $\|\cdot\|_*$ denotes the nuclear norm, given by the sum of singular values, which is a proxy for rank.  
The solution to~\eqref{Eq:RPCA:L1} converges to the solution of~\eqref{Eq:RPCA:L0} with high probability if $\lambda=1/\sqrt{max(n,m)}$, where $n$ and $m$ are the dimensions of $\bX$, given that $\bL$ is low-rank and $\bS$ is sparse. 

The problem in~\eqref{Eq:RPCA:L0} is known as \emph{principal component pursuit} (PCP), and may be solved using the augmented Lagrange multiplier (ALM) algorithm.  
The augmented Lagrangian may be constructed as:
\begin{equation}
\mathcal{L}(\bL,\bS,\bY) = \|\bL\|_* + \lambda\|\bS\|_1 + \langle\bY,\bX-\bL-\bS\rangle + \frac{\mu}{2}\|\bX-\bL-\bS\|_F^2.
\end{equation}
A general solution would solve for the $\bL_k$ and $\bS_k$ that minimize $\mathcal{L}$, update the Lagrange multipliers $\bY_{k+1} = \bY_k + \mu(\bX-\bL_k-\bS_k)$, and iterate until the solution converges.  
For this specific system, the alternating directions method (ADM)~\citep{Lin2010adm,Yuan2009adm} provides a simple procedure to find $\bL$ and $\bS$.  The parameter $\mu$ is discussed more in~\cite{Yuan2009adm} and~\cite{rpca}. 

In the following, RPCA will be used to develop low-dimensional representations for high-dimensional aircraft metrology data (e.g., laser scans or point cloud measurements).  In particular, the left singular vectors $\bPhi$ of the low-rank matrix $\bL$ provide robust principal components, and are computed via the SVD:
\begin{equation}
\bL = \bPhi \mathbf{D} \bV^*.\label{Eq:RPCAmodes}
\end{equation}
These low-rank coherent patterns will then facilitate sparse sensing strategies.

\subsection{Sparse sensor optimization}
Despite the growing abundance of measurement data across the engineering sciences, systems are often fundamentally low-dimensional, exhibiting a few dominant features that may be extracted using dimensionality reduction, such as the RPCA above.  
The existence of low-rank patterns also facilitates sparse sampling, as there are only a few important degrees of freedom that must be characterized, regardless of the ambient measurement dimension.  
In the past decades, tremendous advances have been made for the sparse sampling of low-rank systems.  

Compressed sensing~\citep{donoho2006compressed,candes2006compressive} provides powerful mathematical guarantees for the reconstruction of high-dimensional data that is sparse in a universal encoding basis, such as a Fourier or wavelet basis.  
The resulting algorithms are general, enabling the reconstruction of unknown signals from surprisingly few measurements.  
However, if additional information is available about the structure of the unknown signal (e.g., it is an image of a human face, or a point cloud scan of a particular aircraft part), then the number of measurements required may be further reduced, often dramatically.  
In these cases, a tailored basis may be constructed, for example using PCA or RPCA, and sparse sensors can then be optimized to identify the few relevant coefficients in this basis.  
These methods are generally referred to as \emph{gappy} sampling~\citep{Everson1995gappy,Willcox2006compfl} or empirical interpolation methods (EIM)~\citep{Barrault2004crm,Chaturantabut2010siamjsc,drmac2016siam}, including the discrete empirical interpolation method, or \emph{DEIM}.  
Sparse sampling in a tailored basis has been widely applied to problems in the imaging sciences as well as to develop reduced order models of complex physics, such as unsteady fluid dynamics.  
We briefly summarize these various sparse sampling methods in Sec.~\ref{sec:methods}, following the review of~\cite{Manohar2017csm}.

\section{Methods}
\label{sec:methods}
\begin{figure}
\begin{center}
\begin{overpic}[width=.9\textwidth]{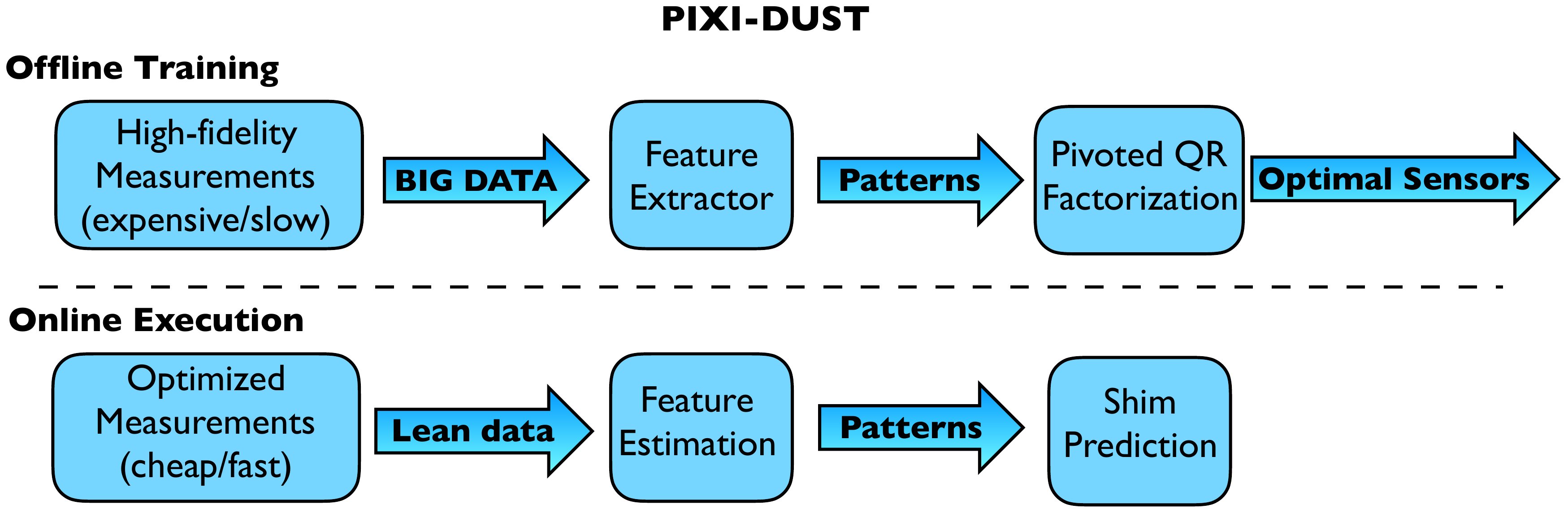}
\end{overpic}
\caption{{\bf Schematic overview of PIXI-DUST method.} Sparse sensor optimization is used to accelerate predictive shimming of aircraft.}\label{FIG:SCHEMATIC}
\end{center}
\end{figure}

In this work, we accelerate the predictive shimming of aircraft by combining sparse sensor optimization with a robust low-dimensional representation of the historical manufacturing data obtained from RPCA.  
The resulting method, termed PIXel Identification Despite Uncertainty in Sensor Technology (PIXI-DUST), is shown schematically in Fig.~\ref{FIG:SCHEMATIC}.  
The term \emph{uncertainty in sensor technology} indicates that the resulting algorithm should be agnostic to the specific metrology technology.  

\subsection{Reconstruction of high-dimensional states from sparse measurements}
High-dimensional scan data is transformed to a low-dimensional representation using the RPCA basis of features $\bPhi_r\in\reals^{n\times r}$, given by the first $r$ left singular vectors of the low-rank matrix $\bL$ in \eqref{Eq:RPCAmodes}: 
\begin{equation}
\reals^n\xrightleftharpoons[~~\bPhi_r^T~~]{\bPhi_r}\reals^r.
\end{equation}
The rank $r$ of the library $\bPhi_r$ is often much smaller than the data dimension $n$. To be explicit, each data instance $\bx\in\reals^n$ may be written
\begin{align}
\label{eqn:PCA_lin_comb}
\bx = \sum_{k=1}^r a_k\bphi_k = \bPhi_r\ba,
\end{align}
where $\ba\in\mathbb{R}^r$, the basis coefficients, provide its low-dimensional representation. Suppose we cannot access the full state $\bx$, but only $p$ point measurements of $\bx$ within the {\em observation} vector $\by\in\reals^n$ 
\begin{align}
\by &= \bC\bx = \bC\bPhi_r\ba \nonumber \\
&= [x_{\ind_1}~x_{\ind_2}\dots x_{\ind_p}]^T,
\end{align}
where $\bC\in\reals^{p\times n}$ is the point measurement operator and $\gamma_i$ are the indices of the measurement locations.
Then, the reconstruction of the remaining state reduces to coefficient estimation via least squares 
\begin{equation}
\label{eqn:ahat}
\hat\ba = (\bC\bPhi)^{\dagger}\by,
\end{equation}
where $\dagger$ is the Moore-Penrose pseudoinverse.  
This procedure, originally known as {\em gappy POD}~\citep{Everson1995gappy}, was first used to reconstruct facial images from a subsampled pixel mask.  Importantly, it permits a drastic reduction in computation by solving for $r\ll n$ unknowns. The full state estimate $\hat\bx$ is subsequently recovered using the feature basis
\begin{equation}
\label{eqn:xhat}
\hat\bx = \bPhi_r\hat\ba.
\end{equation}
The accuracy of reconstruction depends on the structure of the principal components $\bPhi_r$ and the choice of observations $\bC$.  
These features and the choice of measurements are the focus of the next sections.  

\subsection{Features from RPCA}\label{Methods:RPCA}
Shim scan data is collected across multiple aircraft, either leveraging historical data, or collecting data in a streaming fashion as each new aircraft is assembled. 
The shim measurements for a region of interest are flattened into column vectors $\bx_k\in\mathbb{R}^n$, where $n$ corresponds to the number of measurements and $k$ refers to the aircraft line number. 
These vectors are then stacked as columns of a matrix $\bX$:
\begin{equation}
\bX = \begin{bmatrix} \bx_1 & \bx_2 & \cdots & \bx_m\end{bmatrix},
\end{equation}
where the total number of aircraft $m$ is assumed to be significantly smaller than the number of measurements per aircraft, i.e. $m\ll n$. 

Because of tight manufacturing tolerances and a high degree of reproducibility from part to part, it is assumed that the matrix $\bX$ possesses low-rank structure. 
As described above, there may be sparse outliers that will corrupt these coherent features, motivating the use of RPCA to extract the dominant features.  
The data matrix is decomposed as $\bX = \bL + \bS$, and then the low-rank matrix $\bL$ is decomposed into coherent features given by the matrix $\bPhi$, as in~\eqref{Eq:RPCAmodes}.  
We generally use the first $r$ dominant columns of $\bPhi$, corresponding to the $r$ features that explain the most variance in the data.  
The SVD provides the best rank-$r$ least squares approximation for a given rank $r$ by solving the following optimization problem
$$\min_{\hat\bL}\|\bL-\hat\bL\|_2 \mbox{ subject to rank}(\hat\bL) = r. $$
The explicit minimizer of this expression is given by the $r$-truncated SVD of $\hat\bL = \bPhi_r\mathbf{D}_r\bV_r^T$. The target rank $r$ must be carefully chosen so that the selected features only include meaningful patterns and discard noise corresponding to small singular values. 
We accomplish this by determining $r$ according to the optimal singular value truncation threshold of \cite{gavish2014optimal}. 
The subsequent left singular vectors $\bPhi_r$ are the desired low-rank features that span the columns of $\bL$. 
The truncation parameter $r$ ultimately determines the minimal number of observations necessary for a well-posed inverse problem~\eqref{eqn:ahat}.

\subsection{PIXI-DUST sensor selection}
Assume that the sensor measurements are corrupted by  zero-mean i.i.d. Gaussian noise $\bxi\sim N(\mathbf{0},\sigma^2\mathbf{I})$
\begin{equation}
\label{eqn:ynoise}
\by = \bC\bPhi_r\ba + \bxi.
\end{equation}
Then, the error covariance from least squares estimation~\eqref{eqn:ahat} is explicitly defined as
\begin{equation}
\label{eqn:ynoise}
\bSigma = \mbox{Var}(\ba-\hat\ba) = \sigma^2[(\bC\bPhi_r)^T\bC\bPhi_r]^{-1}.
\end{equation}
The features $\bPhi_r$ and noise $\bxi$ are predetermined, but we can choose point measurements $\bC$ to minimize a scalar measure of the ``size" of the error covariance. Thus, sensor selection is cast as an optimization problem over all possible selections to minimize the reconstruction error. 
Intuitively, the best sensors for reconstruction are also most descriptive of the data. 

There are many ways to characterize the ``smallness" of $\bSigma$. For example, point sensors can be chosen to minimize the worst case error variance: either by minimizing the largest eigenvalue of $\bSigma$, or equivalently, maximizing the smallest eigenvalue of $\bSigma^{-1}$,
\begin{equation}
\label{eqn:eopt}
\Max_\bC ~~\lambda_{\min} \left[(\bC\bPhi_r)^T\bC\bPhi_r\right],
\end{equation}
also known as E-optimal experiment design. Alternatively we can minimize the mean square error via the trace of $\bSigma$,
\begin{equation}
\label{eqn:aopt}
\Min_\bC ~~\trace\left[(\bC\bPhi_r)^T\bC\bPhi_r\right]^{-1},
\end{equation}
which is the A-optimal configuration.
We are concerned with overall error reduction along all error components. The error covariance $\bSigma$ characterizes the minimum volume {\em $\eta$-confidence ellipsoid}, $\varepsilon_\eta$, that contains $\ba-\hat\ba$ with probability $\eta$. D-optimal experiment design proposes minimizing the volume of $\varepsilon_\eta$
\begin{equation}
\label{eqn:ellipsoidvol}
\vol(\varepsilon_\eta) = \alpha_{\eta,r}\det\bSigma^{\frac{1}{2}},
\end{equation}
where $\alpha_{\eta,r}$ only depends on $\eta$ and $r$, by minimizing the only term that depends on the sensors: the determinant (hence, {\em D}-optimal). D-optimal sensor optimization is given by:
\begin{equation}
\label{eqn:dopt}
\Max_\bC ~~ \log \det \left[(\bC\bPhi_r)^T\bC\bPhi_r\right].
\end{equation}
Here, we are imposing the following structure on the measurement matrix $\bC\in\reals^{p\times n}$
\begin{equation}
	\bC = [\be_{\gamma_1} ~|~ \be_{\gamma_2} ~|~ \dots ~|~ \be_{\gamma_p}]^T,
\end{equation}
where $\gamma_i\in\{1,\dots,n\}$ indexes the high-dimensional measurement space, and $\be_{\gamma_i}$ are the canonical unit vectors consisting of all zeros except for a unit entry at $\gamma_i$. 
This guarantees that each row of $\bC$ only measures from a single spatial location, corresponding to a point sensor. 
The brute force search over all $n\choose p$ possible sensor selections of $\gamma_i$ in \eqref{eqn:dopt} is combinatorial and computationally intractable even for moderately large $n$ and $p$. A common solution is to relax the binary 0-1 indicators for subset selection to fractional weights over the entire space of measurements. 
In this formulation, the problem is convex and may be solved using standard optimization techniques and semidefinite programs in $O(n^3)$ runtime. Alternatively, greedy methods may be used to select the rows that make sampled features $\bC\bphi_i$ as orthogonal to each other as possible. 
The point sensors $\bC$ are equivalent to row selection along the state dimension of $\bPhi_r$. Since the row selected columns $\bphi_i$ are no longer orthonormal, many greedy row selection methods attempt to maintain near-orthonormality of the features $\bC\bphi_i$ for a well-conditioned inversion. 

We are interested in shrinking all components of the reconstruction error via the associated ellipsoid {\em volume}, which is accomplished using the D-optimal formulation. This work applies a greedy method for matrix volume minimization: the matrix QR factorization of $\bPhi_r$ with column pivoting. The reduced matrix QR factorization with column pivoting decomposes a matrix $\bA\in\reals^{m\times n}$ into a unitary matrix $\bQ$, an upper-triangular matrix $\bR$ and a column permutation matrix $\bC$ such that $\bA\bC^T = \bQ\bR$.

Recall that the determinant of a matrix, when expressed as a product of a unitary factor and an upper-triangular factor, is the product of the diagonal entries in the upper-triangular factor:
\begin{equation}
\left| \det \bA\bC^T \right| = |\det\bQ||\det\bR| = \prod_i |r_{ii}|.
\end{equation}
The pivoted QR alters the matrix $\bPhi_r^T$ with column selection (pivoting) $\bPhi_r^T\bC^T$ to enforce the following diagonal dominance structure on these diagonal entries~\citep{drmac2016siam}:
	\begin{equation}
	\sigma_i^2 = |r_{ii}|^2 \ge \sum_{j=i}^k |r_{jk}|^2; \quad 1\le i \le k \le m.
	\end{equation}
	The column pivoting iteratively increments the volume of the pivoted submatrix by selecting a new pivot column with maximal 2-norm, then subtracting from every other column its orthogonal projection onto the pivot column (see Algorithm \ref{alg:qrpivot}). 	
	In this manner the QR factorization with column pivoting yields $r$ point sensors (pivots or permutation indices) that best characterize the $r$ basis modes $\bPhi_r$
	\begin{equation}
	\bPhi_r^T\bC^T = \bQ\bR.
	\end{equation}
	
	This selects the best sampling points for a well-conditioned inverse in a uniquely determined linear system. We are also interested in the case where there are more measurements than sensors, or $p>r$. Then, the solution is the Moore-Penrose pseudoinverse, the best least-squares regressor to the measurements in an overdetermined system.
	This {\em oversampled} case can be handled using the pivoted QR factorization of $\bPhi_r\bPhi_r^T$, where the column pivots are selected from $n$ candidate measurement indices based on the observation that (denoting $\bTheta = \bC\bPhi_r$)
	\begin{equation}
	\det \bTheta^T\bTheta  = \prod_{i=1}^r \sigma_i(\bTheta\bTheta^T).
	\end{equation}
	Here we drop the absolute value since the determinant of $\bTheta\bTheta^T$ is nonnegative. Accordingly, optimized measurement indices are given by the column permutation matrix $\bC^T$ from the pivoted QR factorization of $\bPhi_r\bPhi_r^T$,
	\begin{equation}
	(\bPhi_r\bPhi_r^T)\bC^T = \bQ\bR.
	\end{equation}
	The QR pivoting algorithm is summarized in Algorithm~\ref{alg:qrpivot}. 
		
	\begin{algorithm*}
		\caption{QR factorization with column pivoting of $\bA\in\reals^{n\times m}$}\label{alg:qrpivot}
		\begin{algorithmic}[1]		
			\Procedure{qrPivot}{ $\bA,~p$ }
			
			\State $\Ind \gets [~~]$
			\For{$k=1,\dots,p$}
			\State $\ind_k = \argmax_{j\notin \Ind} \|\mathbf{a}_j\|_2 $
			\State Find Householder $\tilde{\bQ}$ such that $\tilde{\bQ} \cdot \begin{bmatrix} a_{kk} \\ ~\\ \vdots \\ a_{nk} \end{bmatrix} = \begin{bmatrix}
			\star \\ 0 \\ \vdots \\ 0
			\end{bmatrix}.$  \hfill \mbox{\color{purple}$\star$'s are the diagonal entries of $\bR$}
			\State $ \bA \gets \mbox{diag}(I_{k-1},\tilde{\bQ}) \cdot \bA$ \hfill \mbox{
				\color{purple}
				remove from all columns the orthogonal projection onto $\mathbf{a}_{\ind_k}$}
			\State $\Ind \gets [\Ind,~\ind_k]$
			\EndFor
			\Return $\Ind$
			\EndProcedure
		\end{algorithmic}
	\end{algorithm*}
	
	
	\subsection{Computational considerations}
The pivoted QR algorithm presented in Algorithm~\ref{alg:qrpivot} offers several advantages over competing sensor placement methods. 
In the $p=r$ case, pivoted QR has $O(nr^2)$ runtime, in contrast to standard convex optimization methods that require matrix factorizations of $n\times n$ matrices leading to $O(n^3)$ runtime within each iteration.  
Here, the state dimension $n$ is the limiting factor in the computation since $r\ll n$. 
When $p>r$, our oversampled pivoted QR algorithm requires a single $O(n^3)$ QR factorization and results in a hierarchical list of $n$ optimized pivots, with the first $p$ pivots optimized for sampling $\bPhi_r$ for any $p>r$. 
Alternative optimizations recompute the factorization for each new  $p$, and there are no bounds on the distance of the convex relaxation solution to the optimal sensor selection, due to rounding procedures applied to the fractional weights to obtain the selection. By contrast, QR pivot accuracy is well-studied and upper bounds on the distance to the optimum are derived in~\cite{drmac2016siam}. Their methodology for empirical interpolation, called Q-DEIM, closely resembles our framework but is formulated specifically for reduced order modeling of large-scale dynamical systems.
	
The pivoted QR is a well-established procedure for finding optimal interpolation and quadrature points~\citep{Sommariva2009Fekete,Seshadri2016qr}, and is the  computation underlying the native Matlab backslash operator (`\verb|\|') to solve underdetermined linear systems. 
Similar methods have been employed to discover optimal samples for both linear and nonlinear classification tasks~\citep{Brunton2016siap,Manohar2016jfs,Guo2017sparsetda}. 
Indeed, the pivoted QR is implemented and optimized in most standard scientific computing packages and libraries, including Matlab, LAPACK, NumPy, among others. In addition, QR runtime can be significantly reduced by terminating the procedure after the first $p$ pivots are obtained. The operation can be accelerated further using randomized techniques, for instance, by the random selection of the next pivot~\citep{drmac2016siam} or by using random projections to select blocks of pivots~\citep{Martinsson2017siamjsc,martinsson2015blocked,duersch2015true}. 
	
\vspace{-.075in}

\subsection{PIXI-DUST for shim gaps}
\vspace{-.05in}
Assuming that shim gaps are well characterized by $r$ low-rank features $\bPhi_r$, we seek the optimal $p$ gap measurement locations that best characterize the remaining $n-p$ shim gaps. Subsequently the $n-p$ unknown gaps for a new aircraft are predicted using only $p$ measured gap values at the optimal locations. These optimal locations are obtained in the training stage using the QR column pivoting method above, effectively selecting measurement locations from regions of strong spatial correlation. 

Unknown shim gaps for a new aircraft are predicted (approximated) given only its gap measurements $\by\in\mathbb{R}^r$ at a few optimized locations. Prediction is accomplished with a linear regression on the trained features~\eqref{eqn:xhat}.
We solve for the unknown feature coefficients $\ba$ using only the given gaps at optimal locations in the index set $I$, then predict all remaining gaps using~\eqref{eqn:ahat}, \ie, $\ba = (\bC\bPhi)^{\dagger} \by = \bPhi(I,:)^{\dagger}\by$, and the prediction~\eqref{eqn:xhat}. Finally, we assess the success of the prediction with the percentage of points that differ from the true gap values by less than 0.005", which is the desired measurement tolerance for a laser scanner, or equivalently, the number of indices $i$ out of $n$ total that satisfy $|x_i - \hat x_i | < 0.005$.

Finally, it is useful to decompose a large aircraft structure into several smaller shim components, which are each analyzed independently. 
As we see, this improves the prediction performance and reduces the number of measurements by ensuring that the data remains tightly correlated across aircraft.  

\begin{figure}[t]
\vspace{.1in}
\begin{overpic}[width=0.525\textwidth]{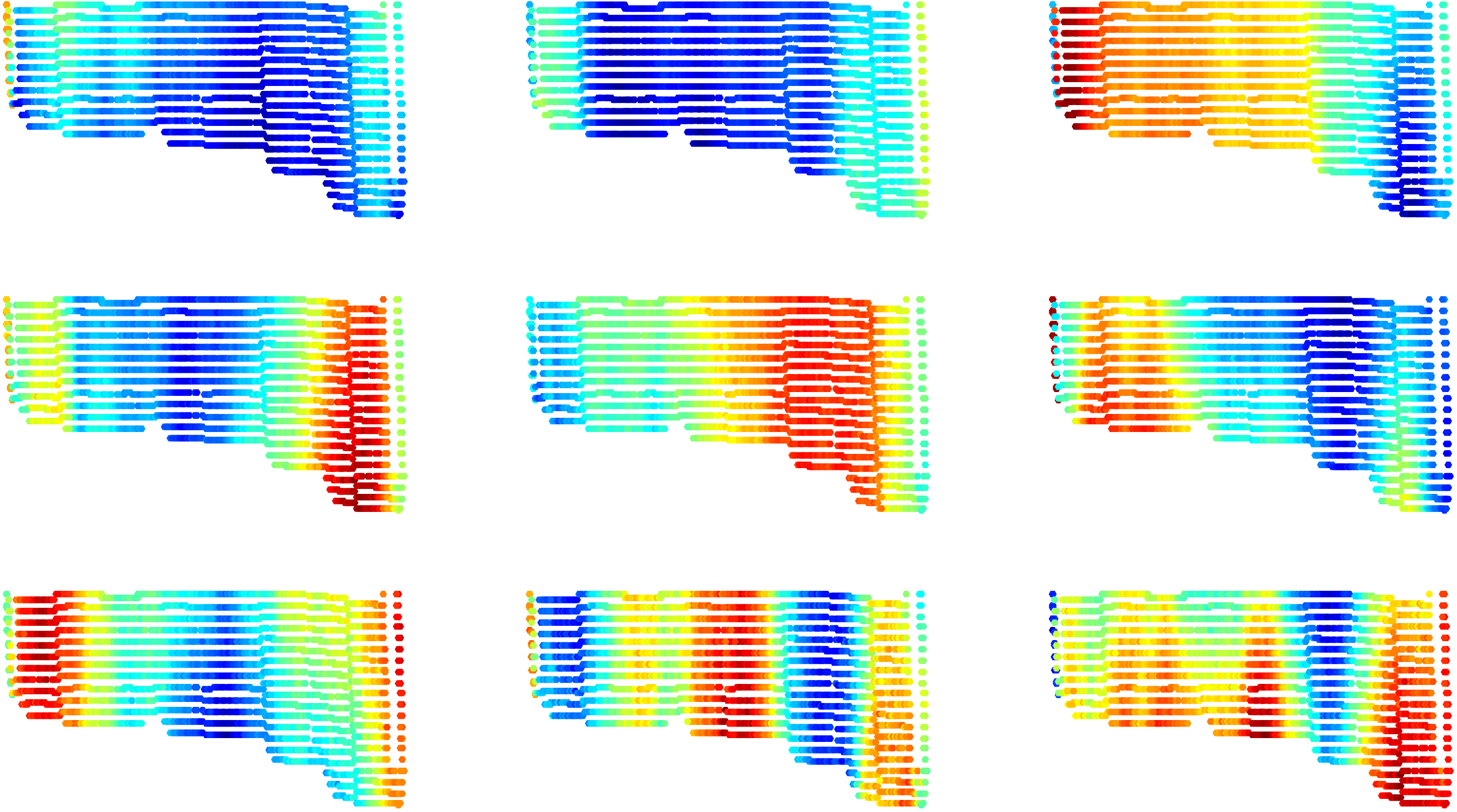} 
\put(-1,62){(a)}
\end{overpic}\quad\quad
	\begin{overpic}[width=0.425\textwidth]{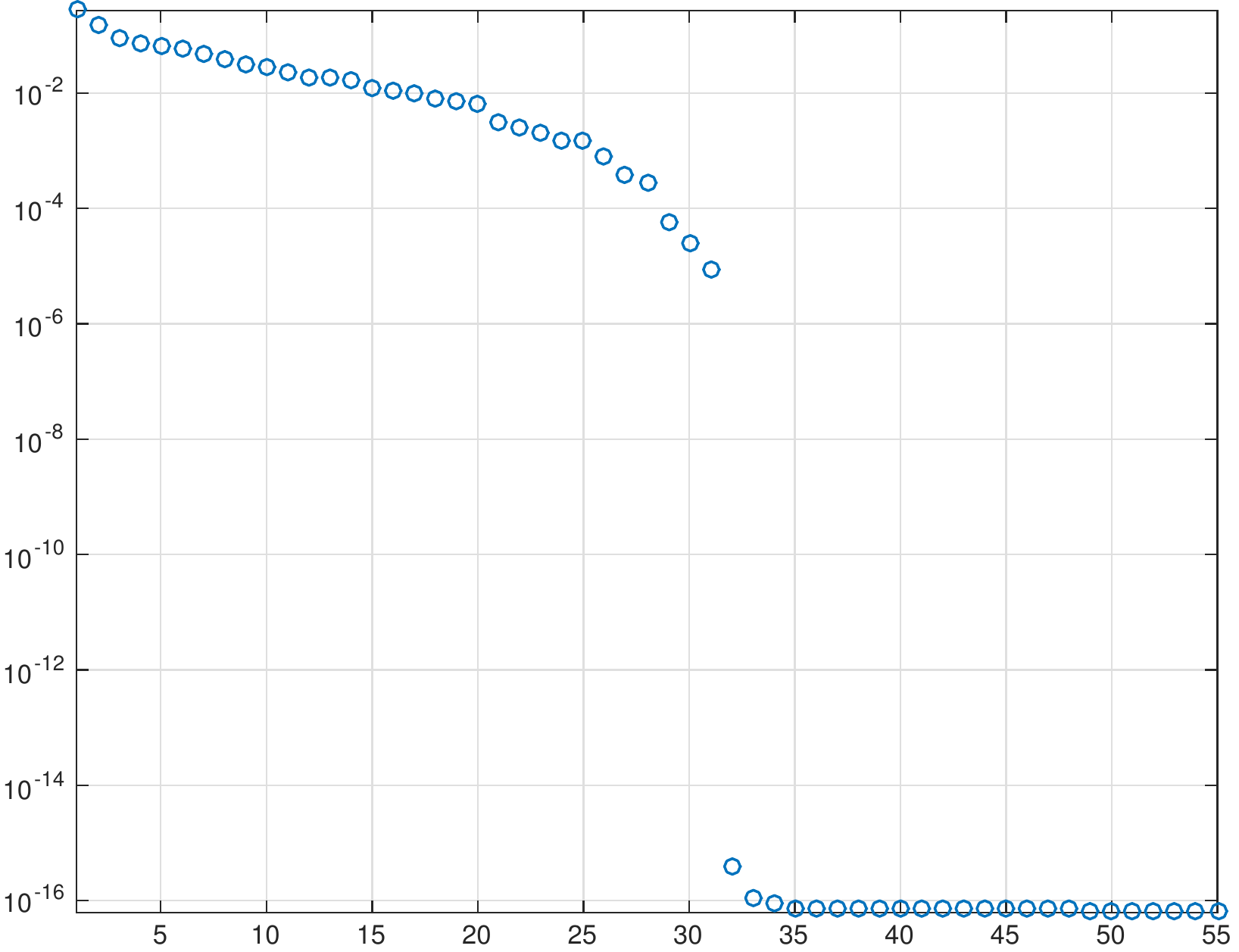}
	\put(-5,76){(b)}
	\put(-5,45){$\sigma_j$}
	\put(50,-4){$j$}
	\end{overpic}
	\vspace{.1in}
	\caption{(a) {\bf RPCA modes 1-9 for shim data.} First nine RPCA features reveal segmented spatial structures reflecting individual shims.  The colormap represents normalized correlations, with largest positive and negative correlations shown in dark red and blue, respectively.  (b) {\bf Normalized singular values for shim data.} Nearly all the variance (energy) in the shim gaps is characterized by the first 31 RPCA features. Thus, the shim dataset intrinsically possesses only 31 degrees of freedom, although each observation is high-dimensional ($n\approx 10000$).\label{fig:sigma_PART}}
\end{figure}

\section{Results}
\label{sec:results}

We demonstrate the performance of the PIXI-DUST architecture on production data from a challenging part-to-part assembly on a Boeing aircraft.  
This data set consists of $10076$ laser gap measurements of the part assembly for each of $54$ aircraft.  
We build a low-order model of the shim distribution using RPCA and then design optimized measurement locations based on these data-driven features.  
We train the model on $53$ aircraft and then validate on the remaining aircraft; this process is repeated for all $54$ possible training/test partitions to cross-validate the results.  
Thus, a data matrix $\bX\in\mathbb{R}^{10076\times 53}$ of training data is constructed, in which each column contains all of the shim gaps for one aircraft, and each row contains the measured gap values at one specific location for all aircraft in the training set. 

The low-rank decomposition of $\bX$ via RPCA is shown in Fig.~\ref{fig:sigma_PART}.  The singular value distribution indicates dominant low-rank structure, and the principal components illustrate the coherent correlated patterns observed in the shim gap data.  
RPCA has a tunable parameter $\lambda$ that controls the strength of outlier rejection or the number and magnitude of nonzero entries in $\bS$. In practice, $\lambda$ is set to an optimal value determined by the dimensions of the matrix, which in our case is $\lambda = {1}/{\sqrt{\max(10076,53)}}$.

\subsection{Prediction results for entire wing to body join}
First, we investigate sensor optimization and shim prediction considering the entire data set as single large structure.  In the next section, we decompose the data into several individual shim segments.  

Based on the PIXI-DUST algorithm, $r=31$ RPCA modes and $p=100$ optimized measurements are identified using oversampled QR pivoting.  
Potential overfitting may occur when fitting more than 100 measurements to these 31 features, since some measurements may be contaminated by noise not reflected in the feature space. 
Thus, increasing the number of measurements further does not yield substantial improvements in prediction accuracy.

\begin{figure}[t]
\vspace{-.1in}
\begin{center}
\includegraphics[width=.6\textwidth]{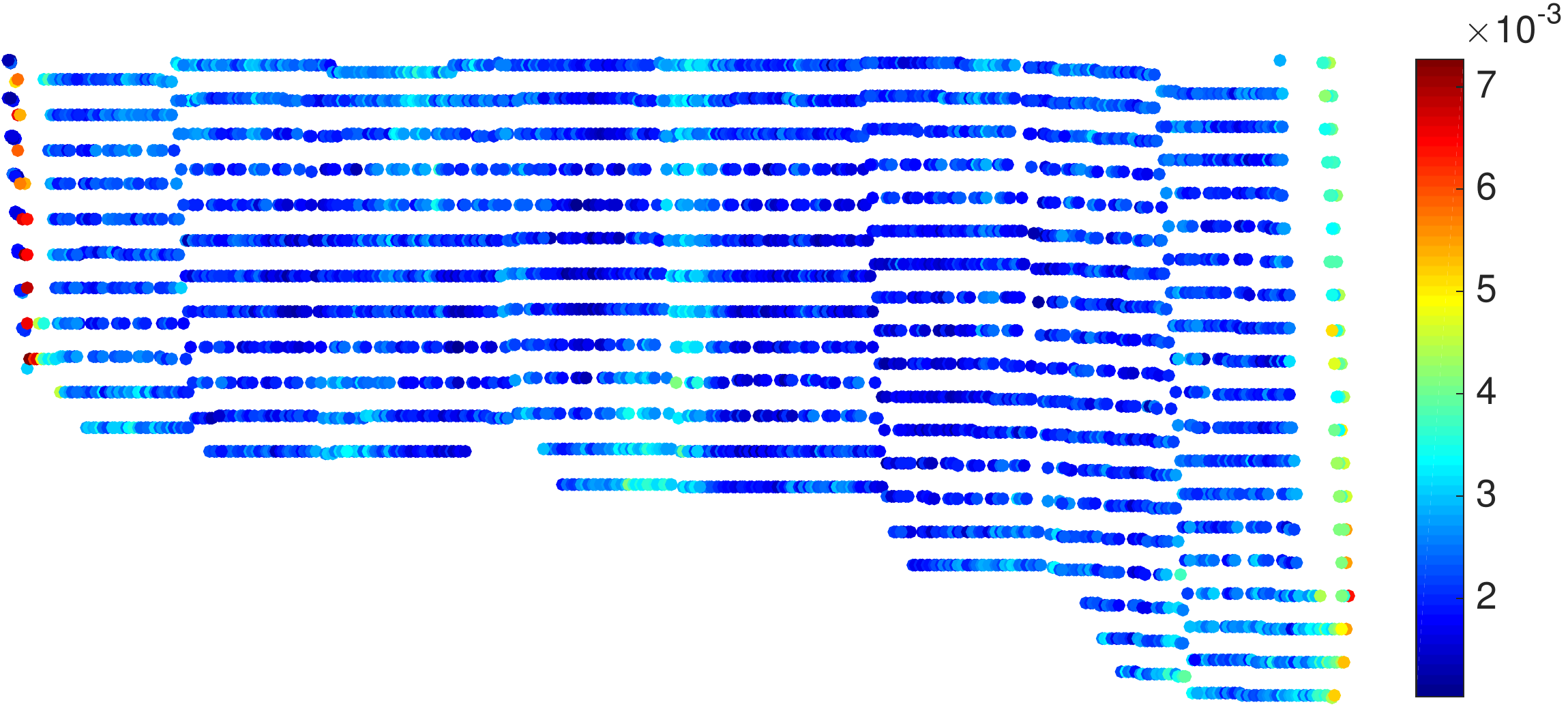}
\vspace{-.1in}
\caption{{\bf Spatial error distribution}. The median pointwise absolute error at most locations is less than the desired measurement tolerance of 0.005 inches. The subsequent figure shows the complete distribution of the reconstruction error. \label{fig:median_PART_error}}
\end{center}
\end{figure}

\begin{figure}[t]
\centering
\begin{subfigure}[t]{.44\textwidth}
	\begin{overpic}[width=1\textwidth]{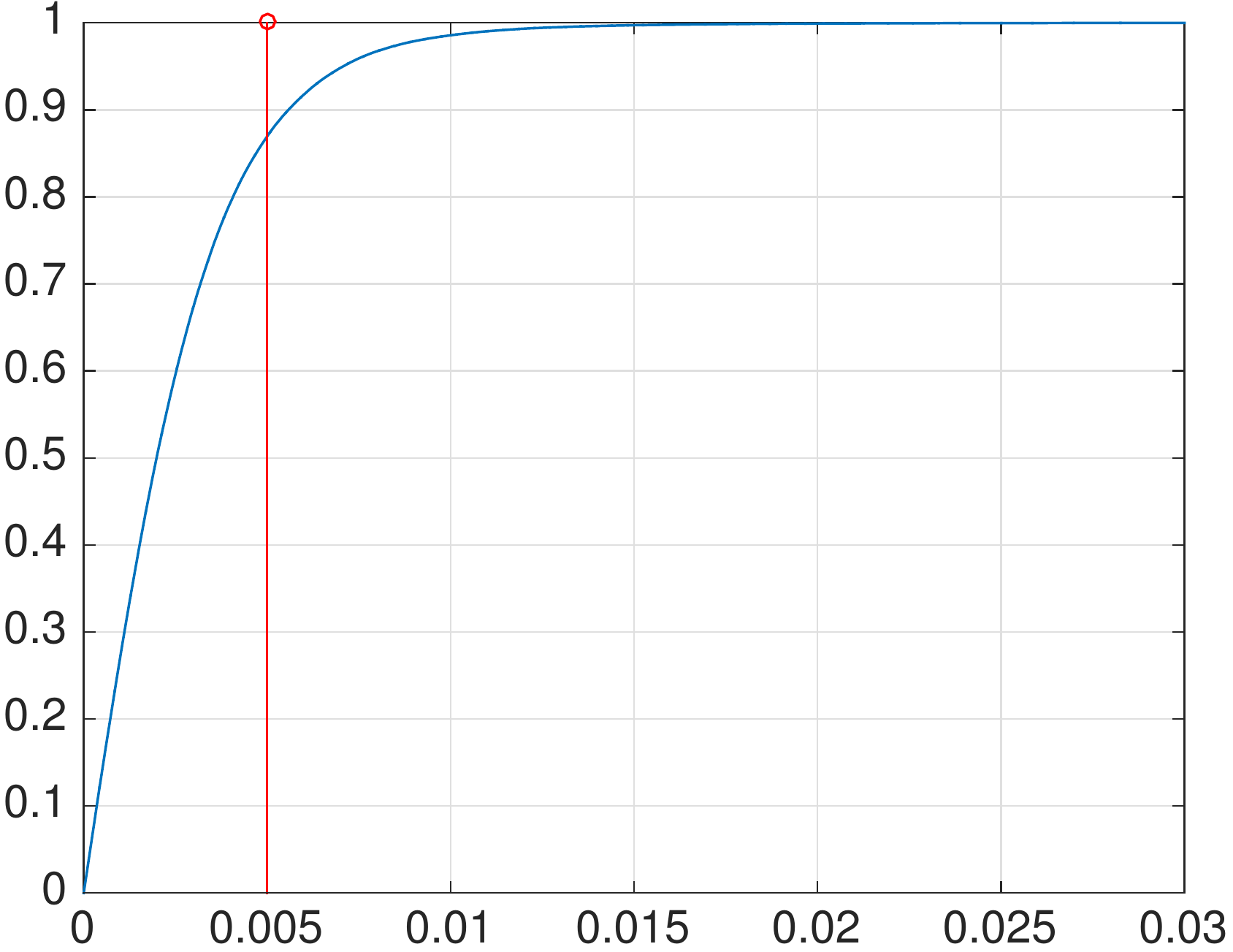}
	\put(35,70) {\small Empirical CDF}
\put(24,-3.7){\small $|x_i-\hat{x}_i|$ Absolute error (in.)}
	\end{overpic}	
\end{subfigure}
\quad
\begin{subfigure}[t]{.46\textwidth}
\begin{overpic}[width=1\textwidth]{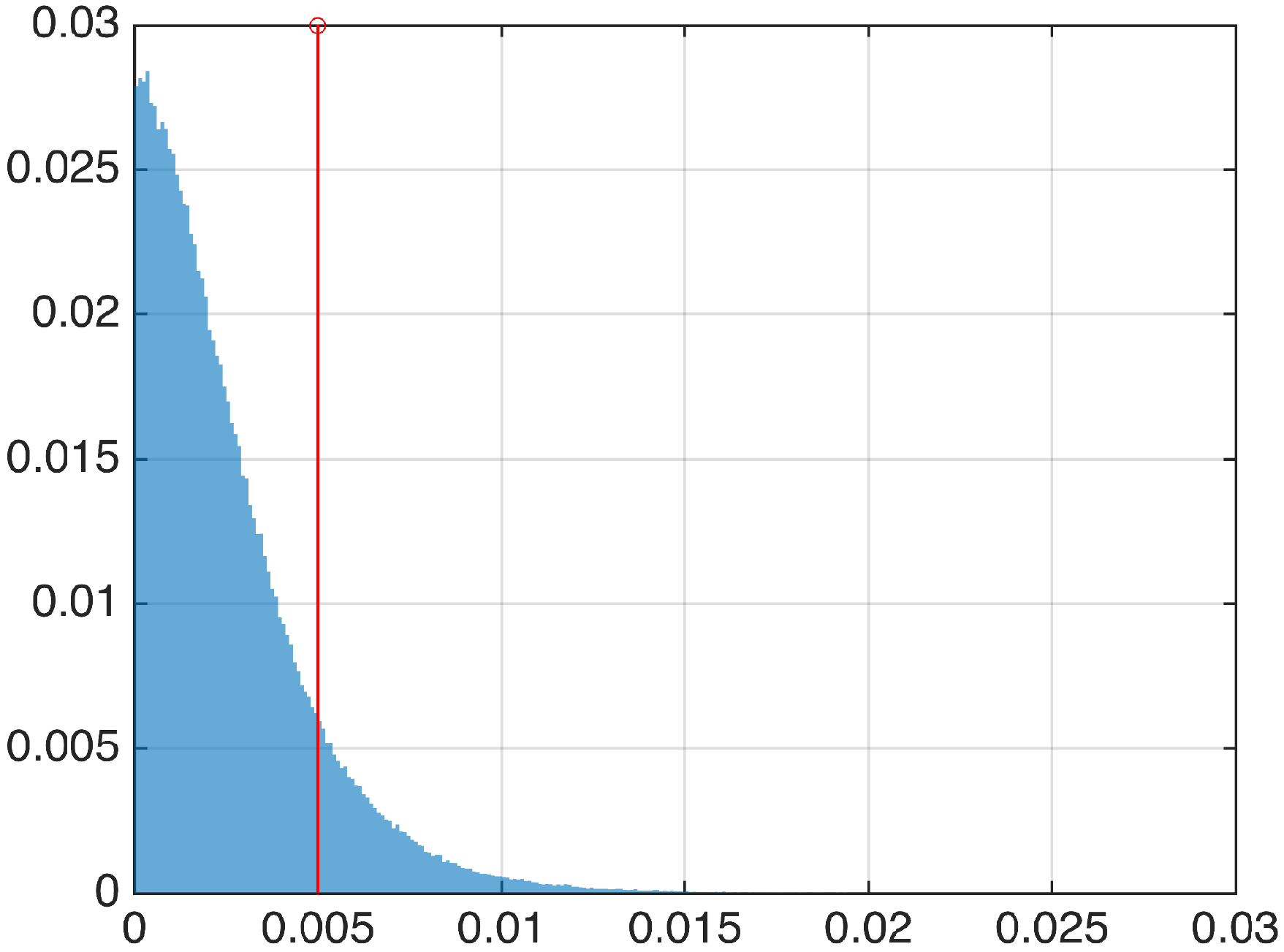}
\put(24,-5){\small $|x_i-\hat{x}_i|$ Absolute error (in.)}
\put(35,67) {\small Prediction error histogram}
\end{overpic}
\end{subfigure}
\vspace{3mm}
\caption{{\bf Pointwise absolute error distribution}. The normalized histogram of absolute error across 534,028 total validation points (10076 gaps $\times$ 53 aircraft) is shown on the right. The cumulative distribution function (CDF) of this distribution (left) shows that 87\% of the predicted gap absolute errors are less than desired tolerance of 0.005 inches, indicated by the red threshold. \label{fig:errordist}}
\end{figure}

The $p=100$ optimized sensors consist of less than $1\%$ of the original $10076$ gap measurements.  
From these few measurements, the remaining shim gaps are predicted with high accuracy.  Figure~\ref{fig:median_PART_error} shows the median pointwise absolute error between the predicted and true shim values. The prediction is successful with high probability across the 53 cross-validation cases. 
The absolute error distribution for all points across all tests, shown in Figure~\ref{fig:errordist}, reveals that 87\% of all points are predicted within $\pm$0.005".  
The few gap locations with error larger than the required $\pm$0.005" will likely not contribute significantly to a manufactured shim, as these are averaged out in splining and fabrication.  
The few points that are outside of this prediction tolerance will be outweighed by the numerous successfully predicted gaps in the fabrication of the shim, as around $1000$ laser scan points are used to define each shim.

\subsection{Prediction results for segmented shims} 

The number of incorrectly predicted gaps can be reduced by training our method on each individual shim part separately, as there are seven individual shim segments in the assembly considered. 
Considering each shim segment separately helps to ensure that data is tightly correlated across aircraft, improving prediction.  
It is possible to reduce the number of total measurements and improve the overall performance using this segmentation approach.  

\begin{figure}
\centering
\begin{subfigure}[t]{.465\textwidth}
\includegraphics[width=1\textwidth]{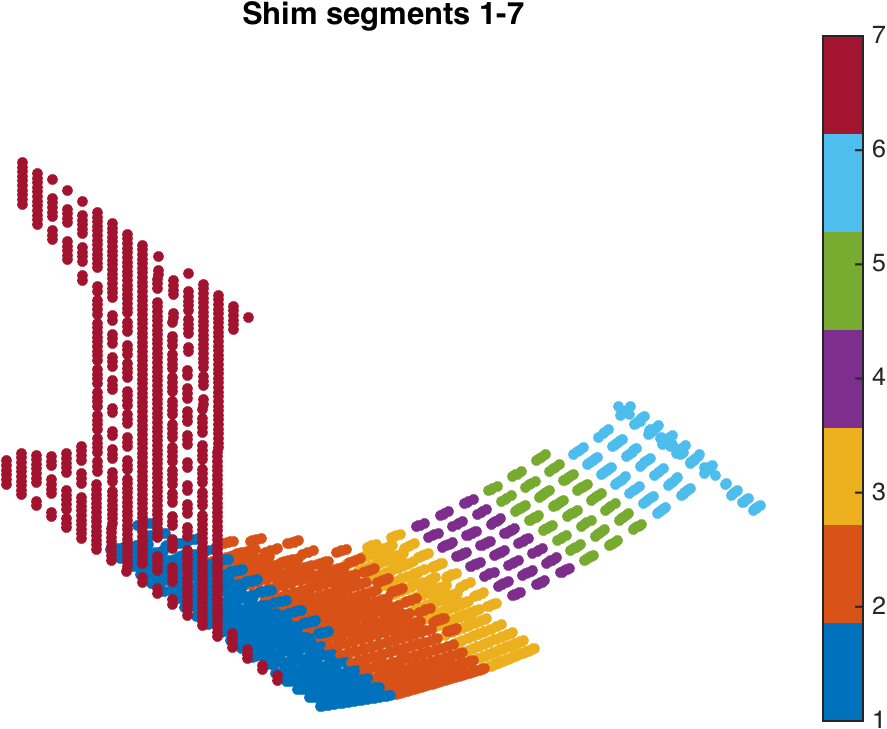}
\end{subfigure}
\quad\quad
\begin{subfigure}[t]{.465\textwidth}
\includegraphics[width=1\textwidth]{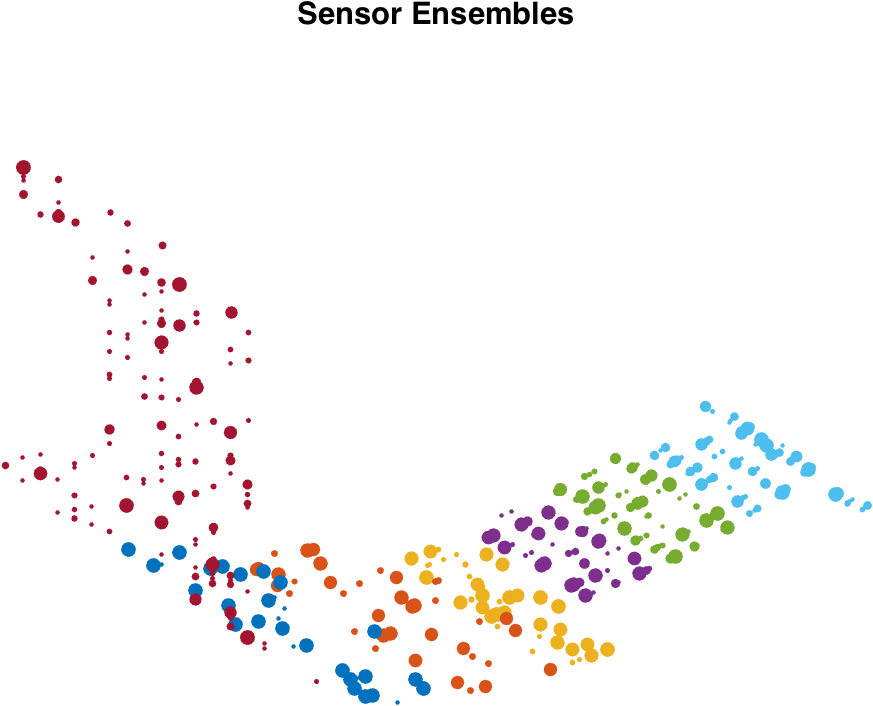}
\end{subfigure}
\caption{{\bf Segmented gap prediction}. Shim boundaries for segmented prediction (left) and sensor ensembles across 53 instances of training (right), sized by the number of times they appear in training.  \label{fig:segmented_pred}}
\end{figure}

Figure~\ref{fig:segmented_pred} displays the seven separately manufactured shim segments, as well as the sensor ensembles for each shim. Prediction results are shown in Table~\ref{Tab:segmented}. Prediction accuracy is vastly improved, and 96-99\% of the shim gap locations are predicted to within the desired $\pm$0.005" tolerance. Furthermore, we note that that the rates of optimal measurements $r$ vary from anywhere between 2\% to 6\% of all points within the shim, which indicates that some shims are higher-dimensional and require more features (hence, sensors) to be fully characterized. This is also reflected in the sensor ensembles in Fig.~\ref{fig:segmented_pred}.

\begin{table*}[b]
	\centering
	\scalebox{1}{
		\begin{tabular}{l l l c c c c c} 
			\hline 			\hline
            \multicolumn{1}{l}{\bf Shim No.}
			& \multicolumn{1}{l}{\bf 1}
			& \multicolumn{1}{c}{\bf 2}
			& \multicolumn{1}{c}{\bf 3}
			& \multicolumn{1}{c}{\bf 4}
			& \multicolumn{1}{c}{\bf 5}            
			& \multicolumn{1}{c}{\bf 6}            
			& \multicolumn{1}{c}{\bf 7}
			\\
			\cmidrule(r){1-8}
			
			\multirow{1}{*}{\rotatebox[origin=c]{0}{\parbox{3cm}{Percent Accurate}}} 
			& 97.90		& 98.05	&  99.82  &  99.94 & 99.99  &  99.03 & 99.97 \\ 					
			\multirow{1}{*}{\rotatebox[origin=c]{0}{\parbox{3.75cm}{Optimal sensors (avg)}}} 
			& 26     &  26	&  25   & 26 &  25  & 26 & 25   \\ 
			\multirow{1}{*}{\rotatebox[origin=c]{0}{\parbox{3cm}{Total points}}} 
			& 1003     &  1116	&  453   & 692 & 709   & 768 &  664   \\ 								%
			\hline \hline
		\end{tabular}
	}
	\caption{Segmented prediction results show vastly improved prediction accuracies, with 97-99\% of gaps predicted to within the desired $0.005$ inch measurement tolerance.}
	\label{Tab:segmented}
\end{table*}

\begin{figure}
	\centering
	\begin{subfigure}[t]{.425\textwidth}
		\includegraphics[width=1\textwidth]{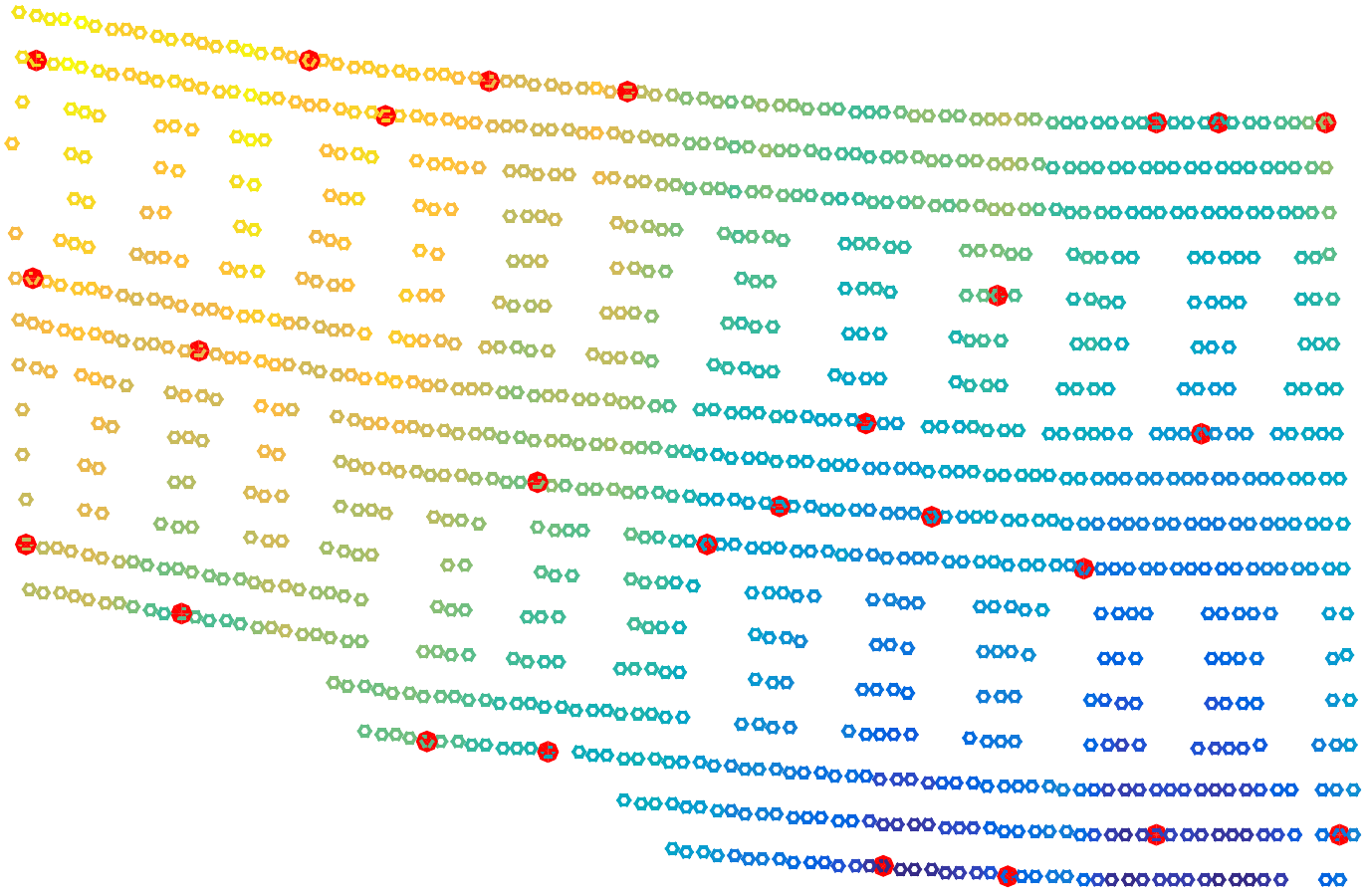}
		\caption{26 sensors}
	\end{subfigure} \quad\quad
	\begin{subfigure}[t]{.425\textwidth}
		\includegraphics[width=1\textwidth]{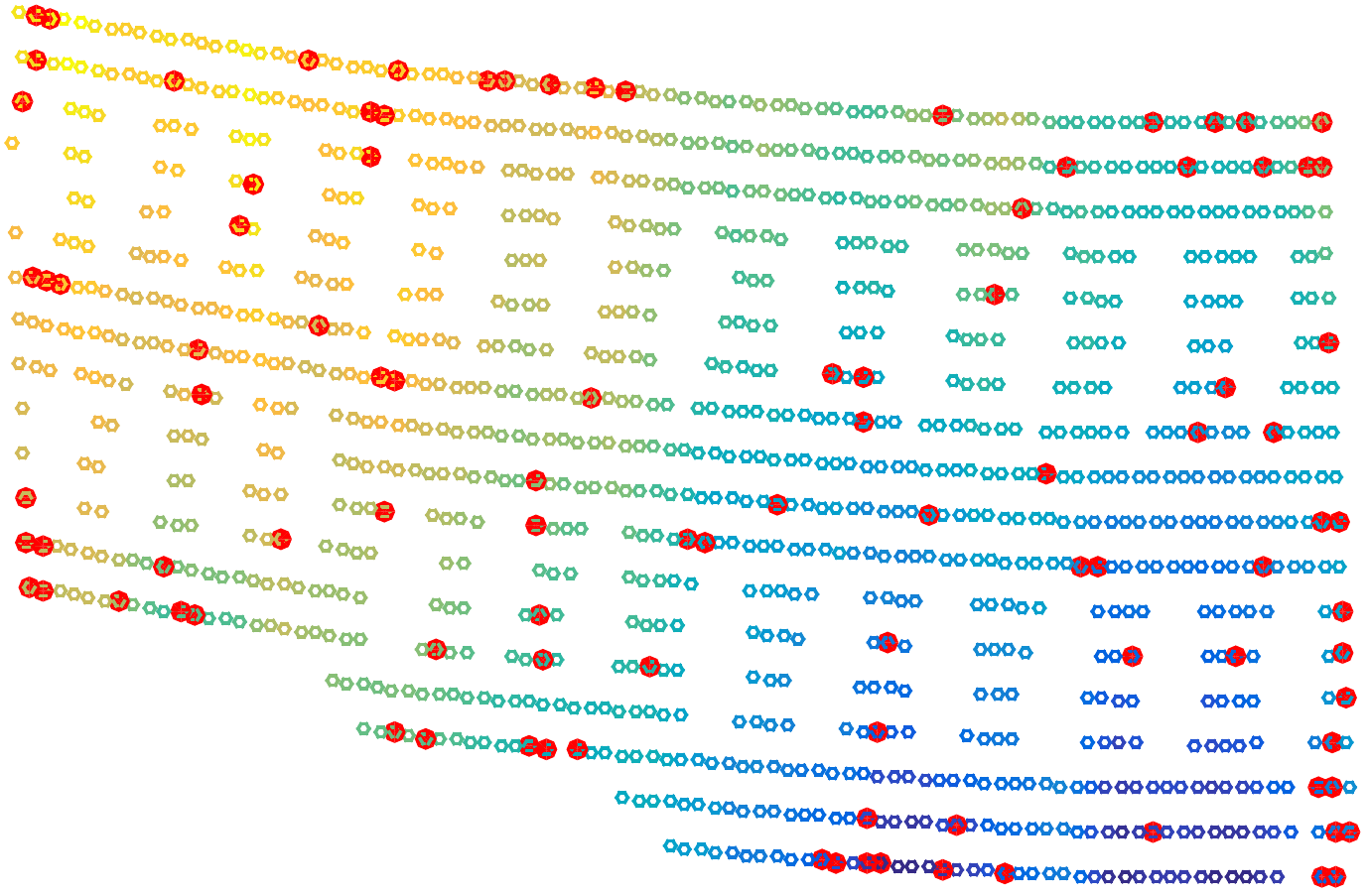}
		\caption{100 sensors}
	\end{subfigure}
	\vspace{-.05in}
	\caption{{\bf Optimized sensors for Shim 2.} The minimal number of sensors is $p=r=26$ since 26 is the intrinsic rank of shim 2 data as computed by the Gavish-Donoho SVD truncation threshold.  Increasing the number of selected sensors increases sensor density at the initial locations, indicating that the data is in fact low-dimensional.  Although additional sensors provide redundant information, it is shown in Fig.~\ref{fig:boxplot_shim2} that they may yield marginal improvements in accuracy. 	\label{fig:shim2sensors}}	
\end{figure}

\begin{figure}
	\centering
	\begin{overpic}[width=.625\textwidth]{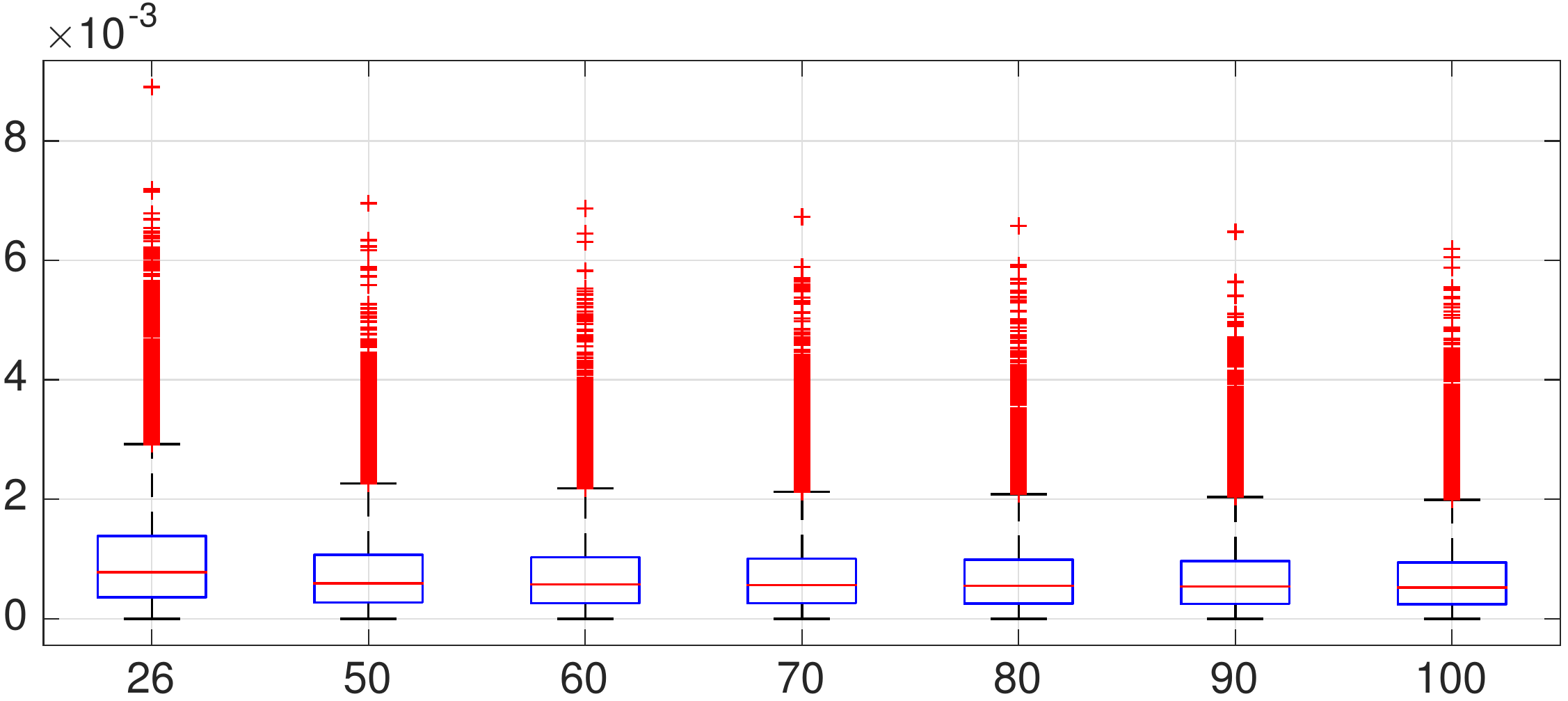}
	\put(36,-4){Number of sensors}
	\put(-5,14){\begin{sideways}Error distribution\end{sideways}}
	\end{overpic}
	\vspace{.2in}
	\caption{{\bf Pointwise error distributions for increasing optimized sensors.} The absolute error distributions do not show much improvement beyond $p=60$ sensors for shim 2, which has 1116 points in total. Furthermore, at $p=80$ only 3 points are mispredicted with absolute error greater than 0.005 inches out of all 1116 $\times$ 53 test points.  \label{fig:boxplot_shim2}}
\end{figure}

\begin{figure}
\begin{center}
\begin{overpic}[width=\textwidth]{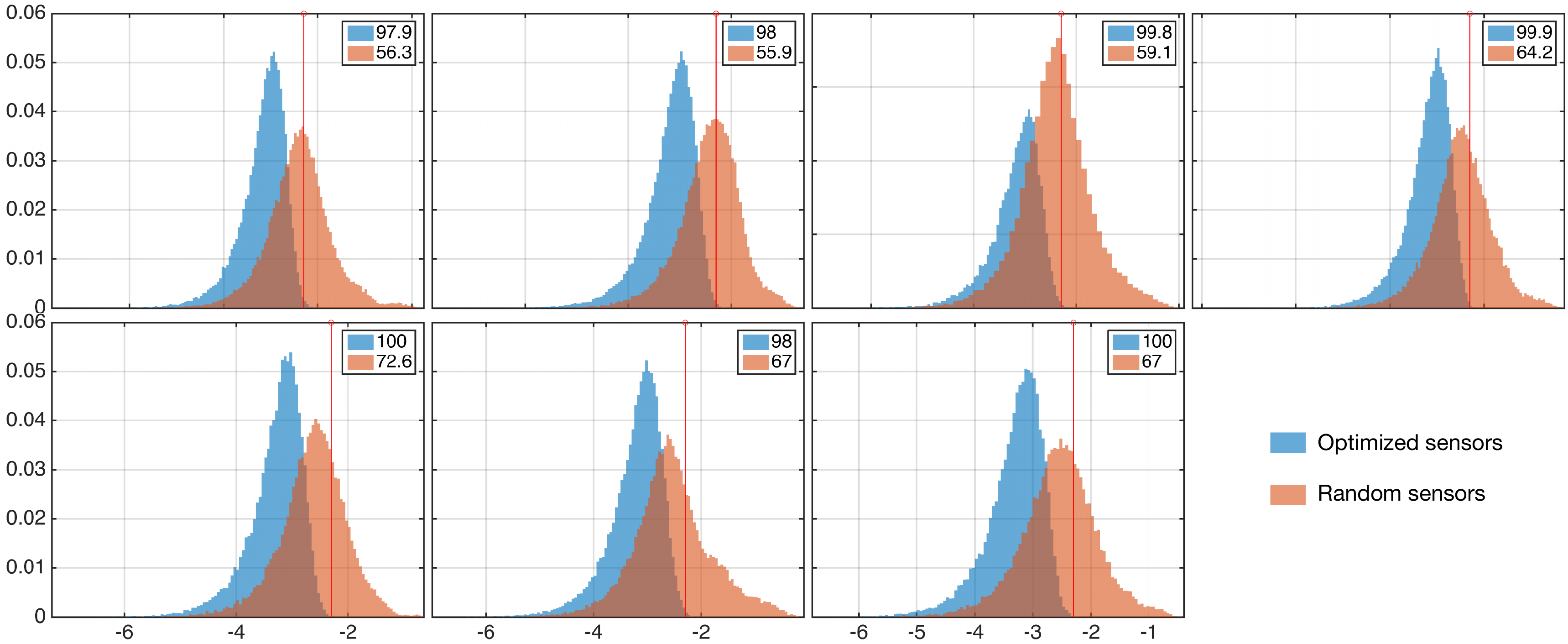}
\put(4,38){Shim 1}
\put(28.5,38){Shim 2}
\put(53,38){Shim 3}
\put(77,38){Shim 4}
\put(4,18){Shim 5}
\put(28.5,18){Shim 6}
\put(53,18){Shim 7}
\put(5,-2){\small  $\log|x_i-\hat{x}_i|$, Error (in.)}
\end{overpic}
\vspace{-.05in}
	\caption{{\bf Absolute error distribution for optimal (blue) vs. random (orange) sensors.} The above are histograms of pointwise absolute error across all validation tests using reconstruction from optimal and random sensors. The red line represents the desired measurement tolerance of 0.005 inches, and the legend indicates the percentage of points that fall within this error tolerance. The histograms indicate that optimal sensors predict nearly twice as well as random sensors. \label{fig:shim_error_hist}}	
\end{center}
\end{figure}
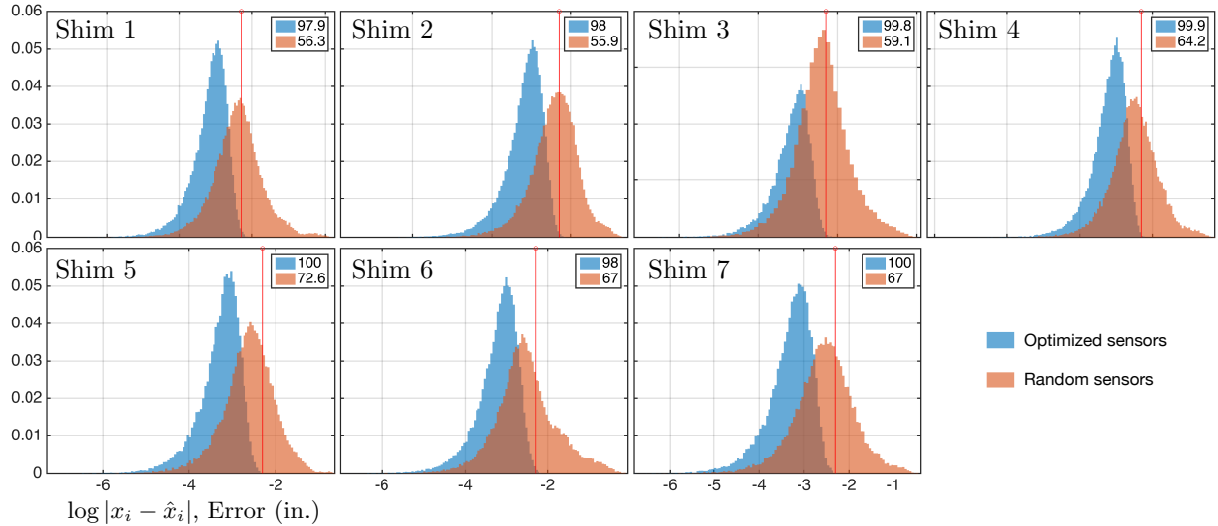

Figure~\ref{fig:shim2sensors} shows the optimized sensor locations for the second shim segment.  As the number of sensors is increased from $p=26$ to $p=100$, the sensor density increases near the originally identified points. 
Figure~\ref{fig:boxplot_shim2} shows the error distribution as the number of sensors is increased, indicating that there is a performance cap at a certain small number of sensors.  
Finally, Fig.~\ref{fig:shim_error_hist} shows the error distributions of predicted shim gaps for all seven segments, comparing the performance of optimized sensors versus randomly chosen measurement locations.  
In all cases, optimized sensors result in significantly more accurate shim prediction over random sensors, with the vast majority of predicted shims within the desired tolerance.

\section{Conclusions and discussion}
\label{sec:conclusions}

This work demonstrates the ability of data-driven sensor optimization to dramatically reduce the number of measurements required to accurately predict shim gaps in aircraft assembly.  
We combine sparse sensor optimization with robust feature extraction and apply the technique to historical Boeing production data from a representative aircraft.  
With around $3\%$ of the original laser scan measurements, our algorithm is able to predict the vast majority of gap values to within the desired measurement tolerance, resulting in accurate shim prediction.  
These optimized measurements exhibit excellent cross-validated performance and may inform targeted, localized laser scans in the future. 
Reducing the burden of data acquisition and downstream computations has tremendous potential to improve the efficiency of predictive shimming applications, which are ubiquitous and often take place in the critical path of aircraft assembly.  
Thus, streamlining this process may result in billions of dollars of savings. 

The proposed PIXI-DUST algorithm has been demonstrated on laser point scan data, although the method is broadly applicable to data from various metrology equipment.  
In the current study, reducing the number of laser scan points directly reduces measurement time, which is considerable. 
However, in many predictive shimming applications, vast and increasing volumes of point cloud data are being collected.  
As opposed to laser scans, point clouds are relatively inexpensive to acquire, and it is less clear that sparse sensing will significantly reduce acquisition time.  
Nonetheless, point cloud scans often yield millions or billions of data points, which result in computationally expensive downstream predictions.  
The sparse sensor algorithm may pinpoint key locations in this vast data, dramatically reducing shim computations, likewise reducing the time required for predictive shimming.  
In addition, because high-resolution scan data is available, it may be possible to use thousands of nearby points in the vicinity of an identified sparse sensor location to obtain a robust gap estimate in this region.  
It may also be possible to obtain \emph{super-resolution}, leveraging the availability of large quantities of nearby data.  

There are a number of interesting future directions for this work.  
In the present study, we use RPCA to remove outliers in the data, providing a robust computation of low-rank patterns. 
However, the sparse outliers may contain information about anomalies and defects, and machine learning  may be used to automatically classify these outliers.  
It has also been observed in this data that aircraft tend to cluster in the PCA feature space, indicating possible differences in manufacturing processes across line number.  
Mining this data and using these PCA coefficients as diagnostics for process control may be possible. 

Although the present work has been applied to historical production data, we are currently working to implement this technology for online learning on future aircraft.  
Importantly, every step in the algorithm is relatively inexpensive, involving laptop computations on the order of minutes.  
The initial RPCA and pivoted QR are one-time computations, which can be accelerated by GPU implementations when datasets contain on the order of millions of points. 
The dimensionality reduction provided by RPCA permit each aircraft's gap data to be compactly represented by its PCA coefficients, plus a small number of PCA features shared by all aircraft. 
Then, the full data can be cheaply recovered from a simply projection onto PCA features. Thus the PCA framework (even without optimal measurements) permits compact storage of large data that often result from scans.  
After the offline sensor optimization, online predictive shimming merely requires 1) collecting a small subset of optimized measurements and 2) performing least-squares regression.  
The features and predictions only improve with increasing data, and so as more aircraft are manufactured, the dominant correlation structures, and resulting sparse sensor locations, become more clear.  
In the first tens of aircraft, full measurements will be collected, and these measurements will then be pared down and optimized as coherent patterns emerge.  
It may also be possible to collect slightly more data than indicated by the optimization to validate predictions.  
When large deviations in predicted shims occur, it may be necessary to temporarily increase sensor resolution or pinpoint changes in the production process.  
It will also be important to investigate how to transfer knowledge acquired on one shim location to other similar parts.  

\subsection*{Acknowledgments}
The authors would like to thank past and present members of the Boeing advanced research center (BARC) predictive shimming team: Kevin Bray, Ron Collins, Matthew Desjardien, Wei Guo, Stephen Holley, Alan Katz, David (Kam To) Law, Seth Pendergrass, and Chace Willcoxson.  
We would also like to thank BARC director Per Reinhall and Boeing leaders Steve Chisolm, Philip Freeman, Thomas Grandine, and Tia Benson Tolle for valuable discussions and strong support.  
Finally, we would like to thank Bing Brunton and Josh Proctor for valuable discussions on sparse sensor technology.  
KM, AB, and SLB acknowledge support from the Boeing Corporation (SSOW-BRT-W0714-0004).  

\bibliographystyle{elsarticle-harv}
\bibliography{papersDimRed}

\begin{thebibliography}{53}
\expandafter\ifx\csname natexlab\endcsname\relax\def\natexlab#1{#1}\fi
\expandafter\ifx\csname url\endcsname\relax
  \def\url#1{\texttt{#1}}\fi
\expandafter\ifx\csname urlprefix\endcsname\relax\def\urlprefix{URL }\fi

\bibitem[{Antolin-Urbaneja et~al.(2016)Antolin-Urbaneja, Livinalli, Puerto,
  Liceaga, Rubio, San-Roman, and Goenaga}]{antolin2016end}
Antolin-Urbaneja, J.~C., Livinalli, J., Puerto, M., Liceaga, M., Rubio, A.,
  San-Roman, A., Goenaga, I., 2016. End-effector for automatic shimming of
  composites. Tech. rep., SAE Technical Paper.

\bibitem[{Barrault et~al.(2004)Barrault, Maday, Nguyen, and
  Patera}]{Barrault2004crm}
Barrault, M., Maday, Y., Nguyen, N.~C., Patera, A.~T., 2004. An `empirical
  interpolation' method: application to efficient reduced-basis discretization
  of partial differential equations. Comptes Rendus Mathematique 339~(9),
  667--672.

\bibitem[{Boyl-Davis et~al.(2014)Boyl-Davis, Jones, and
  Zimmerman}]{boyl2014digitally}
Boyl-Davis, T.~M., Jones, D.~D., Zimmerman, T.~E., Jun.~24 2014. Digitally
  designed shims for joining parts of an assembly. US Patent 8,756,792.

\bibitem[{Boyl-Davis et~al.(2016)Boyl-Davis, Jones, and
  Zimmerman}]{boyl2016methods}
Boyl-Davis, T.~M., Jones, D.~D., Zimmerman, T.~E., Aug.~30 2016. Methods of
  fabricating shims for joining parts. US Patent 9,429,935.

\bibitem[{Brunton et~al.(2016)Brunton, Brunton, Proctor, and
  Kutz}]{Brunton2016siap}
Brunton, B.~W., Brunton, S.~L., Proctor, J.~L., Kutz, J.~N., 2016. Sparse
  sensor placement optimization for classification. SIAM Journal on Applied
  Mathematics 76~(5), 2099--2122.

\bibitem[{Camelio and Hu(2002)}]{camelio2002compliant}
Camelio, J.~A., Hu, S.~J., 2002. Compliant assembly variation analysis using
  components geometric covariance. In: ASME 2002 International Mechanical
  Engineering Congress and Exposition. American Society of Mechanical
  Engineers, pp. 431--437.

\bibitem[{Cand{\`e}s et~al.(2011)Cand{\`e}s, Li, Ma, and Wright}]{rpca}
Cand{\`e}s, E.~J., Li, X., Ma, Y., Wright, J., 2011. Robust principal component
  analysis? Journal of the ACM 58~(3), 11.

\bibitem[{Cand{\`e}s et~al.(2006)}]{candes2006compressive}
Cand{\`e}s, E.~J., et~al., 2006. Compressive sampling. In: Proceedings of the
  International Congress of Mathematicians. Vol.~3. Madrid, Spain, pp.
  1433--1452.

\bibitem[{Carnelio and Yim(2006)}]{carnelio2006identification}
Carnelio, J.~A., Yim, H., 2006. Identification of dimensional variation
  patterns on compliant assemblies. Journal of Manufacturing Systems 25~(2),
  65--76.

\bibitem[{Chaturantabut and Sorensen(2010)}]{Chaturantabut2010siamjsc}
Chaturantabut, S., Sorensen, D.~C., 2010. Nonlinear model reduction via
  discrete empirical interpolation. SIAM Journal on Scientific Computing
  32~(5), 2737--2764.

\bibitem[{Chouvion et~al.(2011)Chouvion, Popov, Ratchev, Mason, and
  Summers}]{chouvion2011interface}
Chouvion, B., Popov, A., Ratchev, S., Mason, C., Summers, M., 2011. Interface
  management in wing-box assembly. Tech. rep., SAE Technical Paper.

\bibitem[{Donoho(2015)}]{Donoho2015data}
Donoho, D., 2015. 50 years of data science. In: Princeton NJ, Tukey Centennial
  Workshop.

\bibitem[{Donoho(2006)}]{donoho2006compressed}
Donoho, D.~L., 2006. Compressed sensing. Information Theory, IEEE Transactions
  on 52~(4), 1289--1306.

\bibitem[{Drmac and Gugercin(2016)}]{drmac2016siam}
Drmac, Z., Gugercin, S., 2016. A new selection operator for the discrete
  empirical interpolation method---improved a priori error bound and
  extensions. SIAM Journal on Scientific Computing 38~(2), A631--A648.

\bibitem[{Duersch and Gu(2015)}]{duersch2015true}
Duersch, J.~A., Gu, M., 2015. True {BLAS}-3 performance {QRCP} using random
  sampling. arXiv preprint arXiv:1509.06820.

\bibitem[{Esmaeilian et~al.(2016)Esmaeilian, Behdad, and
  Wang}]{esmaeilian2016evolution}
Esmaeilian, B., Behdad, S., Wang, B., 2016. The evolution and future of
  manufacturing: A review. Journal of Manufacturing Systems 39, 79--100.

\bibitem[{Everson and Sirovich(1995)}]{Everson1995gappy}
Everson, R., Sirovich, L., 1995. Karhunen--{L}o\`eve procedure for gappy data.
  Journal of the Optical Society of America A 12~(8), 1657--1664.

\bibitem[{Gavish and Donoho(2014)}]{gavish2014optimal}
Gavish, M., Donoho, D.~L., 2014. The optimal hard threshold for singular values
  is 4/sqrt(3). IEEE Transactions on Information Theory 60~(8), 5040--5053.

\bibitem[{Golub and Kahan(1965)}]{Golub1965siamb}
Golub, G., Kahan, W., 1965. Calculating the singular values and pseudo-inverse
  of a matrix. Journal of the Society for Industrial \& Applied Mathematics,
  Series B: Numerical Analysis 2~(2), 205--224.

\bibitem[{Guo and Banerjee(2017)}]{guo2017identification}
Guo, W., Banerjee, A.~G., 2017. Identification of key features using
  topological data analysis for accurate prediction of manufacturing system
  outputs. Journal of Manufacturing Systems 43~(2), 225--234.

\bibitem[{Guo et~al.(2017)Guo, Manohar, Brunton, and
  Banerjee}]{Guo2017sparsetda}
Guo, W., Manohar, K., Brunton, S.~L., Banerjee, A.~G., 2017. Sparse-{TDA}:
  Sparse realization of topological data analysis for multi-way classification.
  arXiv preprint arXiv:1701.03212.

\bibitem[{Harding et~al.(2006)Harding, Shahbaz, Kusiak,
  et~al.}]{harding2006data}
Harding, J., Shahbaz, M., Kusiak, A., et~al., 2006. Data mining in
  manufacturing: a review. Journal of Manufacturing Science and Engineering
  128~(4), 969--976.

\bibitem[{Huber(2002)}]{Huber2002as}
Huber, P.~J., 2002. John w. tukey's contributions to robust statistics. Annals
  of statistics, 1640--1648.

\bibitem[{Jamshidi et~al.(2010)Jamshidi, Kayani, Iravani, Maropoulos, and
  Summers}]{jamshidi2010manufacturing}
Jamshidi, J., Kayani, A., Iravani, P., Maropoulos, P.~G., Summers, M., 2010.
  Manufacturing and assembly automation by integrated metrology systems for
  aircraft wing fabrication. Proceedings of the Institution of Mechanical
  Engineers, Part B: Journal of Engineering Manufacture 224~(1), 25--36.

\bibitem[{LaValle et~al.(2011)LaValle, Lesser, Shockley, Hopkins, and
  Kruschwitz}]{lavalle2011big}
LaValle, S., Lesser, E., Shockley, R., Hopkins, M.~S., Kruschwitz, N., 2011.
  Big data, analytics and the path from insights to value. MIT Sloan Management
  Review 52~(2), 21.

\bibitem[{Lechevalier et~al.(2014)Lechevalier, Narayanan, and
  Rachuri}]{lechevalier2014towards}
Lechevalier, D., Narayanan, A., Rachuri, S., 2014. Towards a domain-specific
  framework for predictive analytics in manufacturing. In: Big Data (Big Data),
  2014 IEEE International Conference on. IEEE, pp. 987--995.

\bibitem[{Lee et~al.(2013)Lee, Lapira, Bagheri, and Kao}]{lee2013recent}
Lee, J., Lapira, E., Bagheri, B., Kao, H.-a., 2013. Recent advances and trends
  in predictive manufacturing systems in big data environment. Manufacturing
  Letters 1~(1), 38--41.

\bibitem[{Lin et~al.(2010)Lin, Chen, and Ma}]{Lin2010adm}
Lin, Z., Chen, M., Ma, Y., 2010. The augmented lagrange multiplier method for
  exact recovery of corrupted low-rank matrices. arXiv preprint
  arXiv:1009.5055.

\bibitem[{Lindau et~al.(2013)Lindau, Lindkvist, Andersson, and
  S{\"o}derberg}]{lindau2013statistical}
Lindau, B., Lindkvist, L., Andersson, A., S{\"o}derberg, R., 2013. Statistical
  shape modeling in virtual assembly using {PCA}-technique. Journal of
  Manufacturing Systems 32~(3), 456--463.

\bibitem[{Lowe et~al.(2007)Lowe, Jayaweera, and Webb}]{lowe2007automated}
Lowe, G., Jayaweera, N., Webb, P., 2007. Automated assembly of fuselage skin
  panels. Assembly Automation 27~(4), 343--355.

\bibitem[{Lu and Zhang(1990)}]{lu1990combined}
Lu, S.~C., Zhang, G., 1990. A combined inductive learning and experimental
  design approach to manufacturing operation planning. Journal of Manufacturing
  Systems 9~(2), 103--115.

\bibitem[{Manohar et~al.(2017{\natexlab{a}})Manohar, Brunton, Kutz, and
  Brunton}]{Manohar2017csm}
Manohar, K., Brunton, B.~W., Kutz, J.~N., Brunton, S.~L., 2017{\natexlab{a}}.
  Data-driven sparse sensor placement. arXiv preprint arXiv:1701.07569. Invited
  for IEEE Control Systems Magazine.

\bibitem[{Manohar et~al.(2017{\natexlab{b}})Manohar, Brunton, and
  Kutz}]{Manohar2016jfs}
Manohar, K., Brunton, S.~L., Kutz, J.~N., 2017{\natexlab{b}}. Environmental
  identification in flight using sparse approximation of wing strain. Journal
  of Fluids and Structures 70, 162--180.

\bibitem[{Maropoulos et~al.(2014)Maropoulos, Muelaner, Summers, and
  Martin}]{maropoulos2014new}
Maropoulos, P., Muelaner, J., Summers, M., Martin, O., 2014. A new paradigm in
  large-scale assembly---research priorities in measurement assisted assembly.
  The International Journal of Advanced Manufacturing Technology 70~(1-4),
  621--633.

\bibitem[{Marsh(2008)}]{marsh2008laser}
Marsh, B.~J., 2008. Laser tracker assisted aircraft machining and assembly.
  Tech. rep., SAE Technical Paper.

\bibitem[{Marsh et~al.(2010)Marsh, Vanderwiel, VanScotter, and
  Thompson}]{marsh2010method}
Marsh, B.~J., Vanderwiel, T., VanScotter, K., Thompson, M., Jul.~13 2010.
  Method for fitting part assemblies. US Patent 7,756,321.

\bibitem[{Martin et~al.(2011)Martin, Muelaner, Wang, Kayani, Tomlinson,
  Maropoulos, and Helgasson}]{martin2011metrology}
Martin, O.~C., Muelaner, J.~E., Wang, Z., Kayani, A., Tomlinson, D.,
  Maropoulos, P.~G., Helgasson, P., 2011. Metrology enhanced tooling for
  aerospace (meta): A live fixturing wing box assembly case study. In: 7th
  International Conference on Digital Enterprise Technology. University of
  Bath, pp. 83--92.

\bibitem[{Martinsson(2015)}]{martinsson2015blocked}
Martinsson, P.-G., 2015. Blocked rank-revealing {QR} factorizations: How
  randomized sampling can be used to avoid single-vector pivoting. arXiv
  preprint arXiv:1505.08115.

\bibitem[{Martinsson et~al.(2017)Martinsson, Quintana~Ort\'{i}, Heavner, and
  van~de Geijn}]{Martinsson2017siamjsc}
Martinsson, P.-G., Quintana~Ort\'{i}, G., Heavner, N., van~de Geijn, R., 2017.
  Householder {QR} factorization with randomization for column pivoting
  ({HQRRP}). SIAM Journal on Scientific Computing 39~(2), C96--C115.

\bibitem[{Milo et~al.(2015)Milo, Roan, and Harris}]{milo2015new}
Milo, M.~W., Roan, M., Harris, B., 2015. A new statistical approach to
  automated quality control in manufacturing processes. Journal of
  Manufacturing Systems 36, 159--167.

\bibitem[{Muelaner et~al.(2013)Muelaner, Martin, and
  Maropoulos}]{muelaner2013achieving}
Muelaner, J., Martin, O., Maropoulos, P., 2013. Achieving low cost and high
  quality aero structure assembly through integrated digital metrology systems.
  Procedia CIRP 7, 688--693.

\bibitem[{Muelaner et~al.(2011)Muelaner, Kayani, Martin, and
  Maropoulos}]{muelaner2011measurement}
Muelaner, J.~E., Kayani, A., Martin, O., Maropoulos, P., 2011. Measurement
  assisted assembly and the roadmap to part-to-part assembly. In: 7th
  international conference on digital enterprise technology. University of
  Bath, pp. 11--19.

\bibitem[{Muelaner and Maropoulos(2011)}]{muelaner2011integrated}
Muelaner, J.~E., Maropoulos, P., 2011. Integrated dimensional variation
  management in the digital factory. In: 7th International Conference on
  Digital Enterprise Technology. University of Bath, pp. 39--46.

\bibitem[{Muelaner and Maropoulos(2010)}]{muelaner2010design}
Muelaner, J.~E., Maropoulos, P.~G., 2010. Design for measurement assisted
  determinate assembly ({MADA}) of large composite structures. In: Journal of
  the Coordinate Metrology Systems Conference. University of Bath.

\bibitem[{Ren et~al.(2017)Ren, Cui, Sun, and Cheng}]{ren2017multi}
Ren, L., Cui, J., Sun, Y., Cheng, X., 2017. Multi-bearing remaining useful life
  collaborative prediction: A deep learning approach. Journal of Manufacturing
  Systems 43~(2), 248--256.

\bibitem[{Seshadri et~al.(2016)Seshadri, Narayan, and
  Mahadevan}]{Seshadri2016qr}
Seshadri, P., Narayan, A., Mahadevan, S., 2016. Optimal quadrature subsampling
  for least squares polynomial approximations. arXiv preprint arXiv:1601.05470.

\bibitem[{Sharma et~al.(2000)Sharma, Rajagopal, and Anand}]{sharma2000genetic}
Sharma, R., Rajagopal, K., Anand, S., 2000. A genetic algorithm based approach
  for robust evaluation of form tolerances. Journal of Manufacturing Systems
  19~(1), 46.

\bibitem[{Sommariva and Vianello(2009)}]{Sommariva2009Fekete}
Sommariva, A., Vianello, M., 2009. Computing approximate {F}ekete points by
  {QR} factorizations of {V}andermonde matrices. Computers \& Mathematics with
  Applications 57~(8), 1324--1336.

\bibitem[{Valenzuela et~al.(2015)Valenzuela, Boyl-Davis, and
  Jones}]{valenzuela2015systems}
Valenzuela, D., Boyl-Davis, T.~M., Jones, D.~D., Jan.~21 2015. Systems,
  methods, and apparatus for automated predictive shimming for large
  structures. US Patent App. 14/601,600.

\bibitem[{Vasquez et~al.(2014)Vasquez, Boyl-Davis, Valenzuela, and
  Jones}]{vasquez2014systems}
Vasquez, C.~M., Boyl-Davis, T.~M., Valenzuela, D.~I., Jones, D.~D., Apr.~29
  2014. Systems and methods for robotic measurement of parts. US Patent App.
  14/265,212.

\bibitem[{Willcox(2006)}]{Willcox2006compfl}
Willcox, K., 2006. Unsteady flow sensing and estimation via the gappy proper
  orthogonal decomposition. Computers \& Fluids 35~(2), 208--226.

\bibitem[{Yuan and Yang(2009)}]{Yuan2009adm}
Yuan, X., Yang, J., 2009. Sparse and low-rank matrix decomposition via
  alternating direction methods. preprint 12.

\bibitem[{Zhang et~al.(2006)Zhang, Zhu, Djurdjanovic, and
  Ni}]{zhang2006comparative}
Zhang, M., Zhu, J., Djurdjanovic, D., Ni, J., 2006. A comparative study on the
  classification of engineering surfaces with dimension reduction and
  coefficient shrinkage methods. Journal of Manufacturing Systems 25~(3),
  209--220.

\end{thebibliography}
\end{document}